\documentclass[letterpaper]{article} 
\usepackage{aaai24}  
\usepackage{times}  
\usepackage{helvet}  
\usepackage{courier}  
\usepackage[hyphens]{url}  
\usepackage{graphicx} 
\usepackage{natbib}
\usepackage{caption} 
\usepackage{algorithm}
\usepackage{amsmath}
\usepackage{algorithmic}
\usepackage{amsfonts}
\usepackage{threeparttable}
\usepackage{multirow}
\usepackage{subfigure}
\usepackage{booktabs}
\usepackage{diagbox}
\usepackage{appendix}
\usepackage{newfloat}
\usepackage{listings}
\DeclareCaptionStyle{ruled}{labelfont=normalfont,labelsep=colon,strut=off} 
\floatstyle{ruled}
\newfloat{listing}{tb}{lst}{}
\floatname{listing}{Listing}
\usepackage{listings}
\title{Rethinking Mutual Information for Language Conditioned Skill Discovery on Imitation Learning}
\author {
    Zhaoxun Ju\textsuperscript{\rm 1},
    Chao Yang\textsuperscript{\rm 2},
    Hongbo Wang\thanks{Correspondence to Hongbo Wang, Fuchun Sun.}\textsuperscript{\rm 1},
    Yu Qiao\textsuperscript{\rm 2}
    Fuchun Sun\footnotemark[1]\textsuperscript{\rm 1,\rm 3},
}
\affiliations {
    \textsuperscript{\rm 1}Academy for Engineering and Technology, Fudan University\\
    \textsuperscript{\rm 2}Shanghai Artificial Intelligence Laboratory\\
    \textsuperscript{\rm 3}Department of Computer Science and Technology,
Tsinghua University\\
    zxju21@m.fudan.edu.cn, \{yangchao,qiaoyu\}@pjlab.org.cn, Wanghongbo@fudan.edu.cn, fcsun@tsinghua.edu.cn
}
\usepackage{bibentry}

\begin{document}

\maketitle

\begin{abstract}
Language-conditioned robot behavior plays a vital role in executing complex tasks by associating human commands or instructions with perception and actions. The ability to compose long-horizon tasks based on unconstrained language instructions necessitates the acquisition of a diverse set of general-purpose skills.
However, acquiring inherent primitive skills in a coupled and long-horizon environment without external rewards or human supervision presents significant challenges. In this paper, we evaluate the relationship between skills and language instructions from a mathematical perspective, employing two forms of mutual information within the framework of language-conditioned policy learning.
To maximize the mutual information between language and skills in an unsupervised manner, we propose an end-to-end imitation learning approach known as Language Conditioned Skill Discovery (LCSD). Specifically, we utilize vector quantization to learn discrete latent skills and leverage skill sequences of trajectories to reconstruct high-level semantic instructions.
Through extensive experiments on language-conditioned robotic navigation and manipulation tasks, encompassing BabyAI, LORel, and CALVIN, we demonstrate the superiority of our method over prior works. Our approach exhibits enhanced generalization capabilities towards unseen tasks, improved skill interpretability, and notably higher rates of task completion success.
\end{abstract}

\section{Introduction}
General-purpose robots operating alongside humans in their environment must develop the ability to understand and respond to human language in order to perform a wide range of complex tasks. Currently, there is significant research interest in language-conditioned policy learning methods, such as Vision-Language Navigation (VLN) \cite{gu2022vision} and Vision-Language Manipulation (VLM) \cite{hiveformer, peract}, which aim to enable robots to learn the connection between language instructions and their perceptions and actions.

In multi-task scenarios, tasks are typically defined by different task IDs \cite{kitchen,metaworld}. However, in complex environments, task IDs do not capture the relationships between tasks effectively and can be labor-intensive to define. On the other hand, human language provides a more natural and flexible way to define and specify tasks. Additionally, robots need to acquire a diverse set of general-purpose skills that enable them to understand unconstrained language instructions and perform long-horizon tasks.

Most modern skill-learning methods are limited to task ID settings and sparse reward reinforcement learning (RL) environments. Hierarchical reinforcement learning (HRL) approaches to address complex tasks by learning latent skills, which are then used in low-level meta-control \cite{HRL_Latent}. 
Other approaches decouple skill state mutual information into forward \cite{dads,EDL,CIC} and reverse \cite{VIC,DIYAN,reverse} forms, which are incorporated into the reward function. These works offer theoretical analysis and outperform other methods in RL benchmarks \cite{mujoco}. However, these approaches have not been applied to language-conditioned policies.

\begin{figure}[H]
	\centering
    \includegraphics[width=1\linewidth]{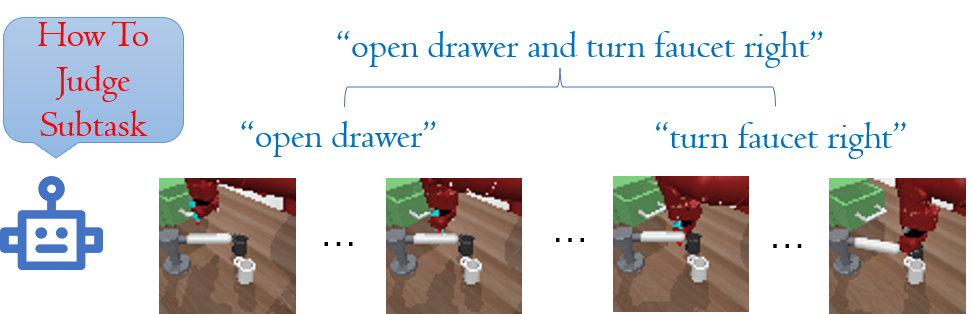}
    \caption{\textbf{An example of multi-task language conditioned situation.} When confronted with intricate language instructions such as "open the drawer and turn the faucet to the right," the agent must decipher and execute the tasks based on the current state.}
    \label{fig:fisrt}
\end{figure}

As depicted in Figure \ref{fig:fisrt}, given a task specification like \textit{open drawer and turn faucet right}, traditional language-conditioned policy struggles to effectively differentiate the subtasks contained within language instructions based on different states \cite{hiveformer}. Contrastive learning is commonly employed for establishing multimodal relationships \cite{contrastive}. However, this approach typically requires pre-labeling of corresponding image sequences and language subtasks, which can hinder generalization. By learning discrete skills, we can fully demonstrate the generalization ability of our imitation model in multi-task scenarios without refining tasks.

Mapping complex languages to discrete skill spaces presents a challenge. In this paper, we experimentally found that skills can directly relate to language instructions, allowing for direct optimization based on their mutual relation. Moreover, in multi-task language-conditioned environments, as illustrated in Figure \ref{fig:fisrt}, latent skills specified in language instructions need to be constrained by the state.

To address these challenges, we propose the Language Conditioned Skill Discovery (LCSD) method to tackle the imitation learning problem in multi-task environments. Our approach is based on mutual information theory, which establishes the relationship between discrete skills, the current state, and language instructions. We employ the VQ-VAE method \cite{vqvae} for skill learning, where the encoder decomposes language and the current state while the decoder aims to reconstruct unique discrete skills and convert them back into language. To generate diverse skills, we introduce code reinitialization to prevent index collapse. We utilize the diffusion policy with the U-net denoising model as an imitation policy, which exhibits better adaptability to different environments.

We conduct experiments in robotic manipulation and 2D navigation to evaluate the effectiveness of LCSD. Compared with language condition policies and skill-based imitation models, our method outperforms prior works. LCSD demonstrates superior generalization, skill interpretability, and task completion rates. Notably, it achieves a 20\% improvement in complex robot manipulation tasks.

To summarize, our contributions are as follows:
\begin{itemize}

\item  We propose a skill-learning method based on mutual information that establishes the relationship between state, skill, and language.

\item  We introduce LCSD, a hierarchical skill learning Imitation policy based on VQ-VAE and diffusion model for long-horizon, language-conditioned multi-task environments.

\item We show that our skill discovery method provides better interpretable discrete skills in different environmental conditions than previous methods.

\item  We demonstrate that our method outperforms existing methods in language-conditioned multi-task environments.


\end{itemize}

\section{Related Work}
\subsection{Language Conditioned Policy}  
Prior research has primarily addressed decision-making in complex tasks that involve language instructions, particularly in robot environments \cite{cliport,LORel}. Existing work has focused on employing pre-trained language models \cite{clip,bert,palm} as language encoders due to the complexity and diversity of human languages. Some previous studies have used behavior cloning to align the output of pre-trained language models with observation inputs in order to predict actions \cite{cliport,vlmbench}. Other approaches have explored LLM (Large Language Model) prompt engineering to decompose complex language instructions into sub-tasks \cite{languageprompt,saycan}. A closely related work to ours is Saycan \cite{saycan}, as both our work and Saycan aim to generalize latent skills using languages and states. However, Saycan requires a pre-defined set of skills to estimate the Q-function for each skill, whereas we can extend our skills to unknown tasks by utilizing a codebook of varying sizes.

\subsection{Skill Discovery via mutual information}
Skill discovery has been primarily employed in Hierarchical Reinforcement Learning (HRL). Agents select latent variables from a set of skills at the high-level policy, which is then executed by a meta-controller to perform sub-tasks \cite{HRL_Latent,skimo}. Recent studies have emphasized encouraging agents to explore and have often relied on the mutual information between states and skills \cite{VIC,EDL}. However, few works have addressed skill learning in a language-conditioned environment. LISA \cite{lisa} utilizes a skill predictor based on states and language within specific horizons. Nevertheless, a single encoder cannot establish a direct connection between skills and language, leading to instability in skill learning.


\section{Preliminary}

\label{sec:pre}

\textbf{Mutual Information Skill learning:} Mutual Information(MI) is a measure of the statistical dependence between two variables. Given state $s$ and skill $z$, the mutual information $I(z;s)$ can be optimized in two ways \cite{EDL}. The forward form: $I(z;s)=H(s)-H(s|z)$, where $p(s|z)$ is estimated by a variational approximation, state entropy is approximated by expectations of $p(s|z)$ estimated over all skills \cite{EDL,dads,CSD}. In the reverse form $I(z;s)=H(z)-H(z|s)$, latent code $z$ is sampled from a fixed distribution and the lower bound of conditioned entropy is estimated by $\rho_\pi(z|s)$ \cite{DIYAN,VIC}.

\begin{figure*}
  \centering
  \includegraphics[width=1\linewidth]{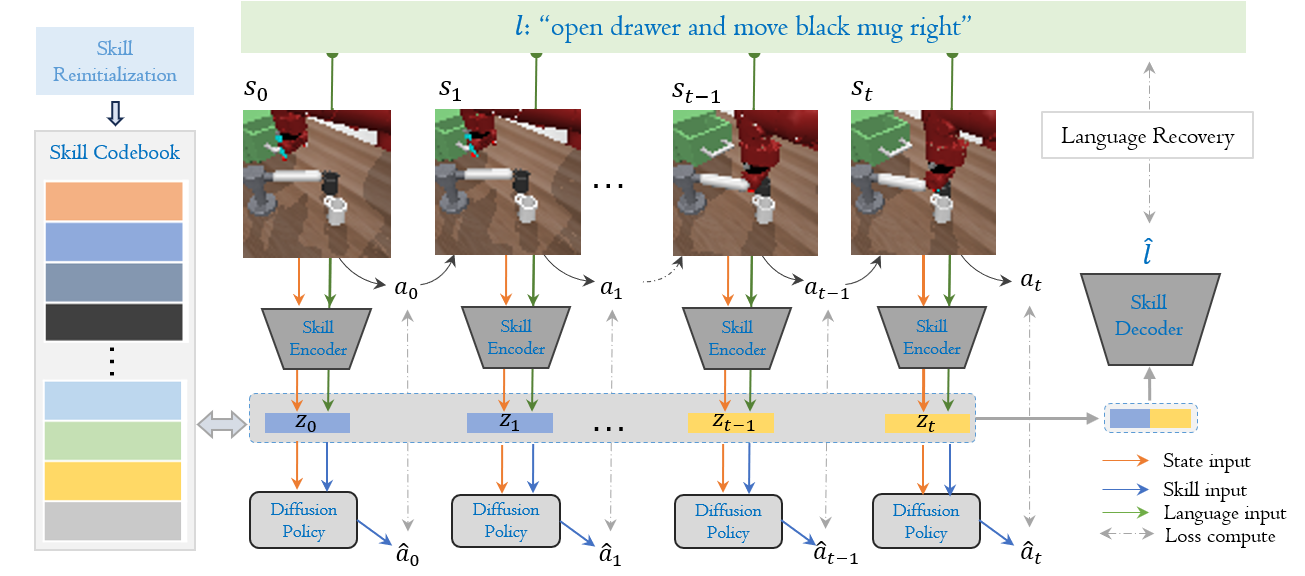}
  \caption{\textbf{Overview of LCSD}. In the skill learning stage, the encoder decomposes the current state and language to a lower-dimensional latent space, while the decoder recovers the quantized latent skills to the language embeddings. A single vector is chosen from the codebook in each step and used to quantize the encoder outputs. The diffusion model is used as an action predictor conditioning on current state and skill(or language). }
  \label{fig:overall}
\end{figure*}
\textbf{VQ-VAE:} Vector Quantized Variational Autoencoder (VQ-VAE) \cite{vqvae} is a neural network architecture for unsupervised learning of latent representations of data. In VQ-VAE, the encoder maps the input data to a continuous latent space, which is then quantized to a discrete codebook. The decoder maps the discrete code to the output space, generating new samples. VQ-VAE updates the encoder, decoder, and codebook parameters with the following loss function.
\begin{equation}
\begin{split}
    L=\log p\left(x \mid q(z_q^k)\right)+\left\|s g\left[p(x)\right]-z_q^k\right\|_2^2+ \\ 
    \beta\left\|p(x)-s g[z_q^k]\right\|_2^2
\end{split}
\label{eq:VQ-VAE}
\end{equation}
the first term represents the reconstruction from discrete code to original input for updating the encoder $p$ and decoder $q$. The second term leads the discrete vectors in the codebook $z_q^{1...N}$ to approach the nearest output of the encoder, while the last term is commitment loss, encouraging the output of the encoder to stay close to the chosen kth codebook vector $z_q^k$. The codebook update can also use exponential moving averages instead of the second in the loss function.

\section{Approach}
\label{sec:approach}
LCSD is a two-stage imitation learning structure that comprises an encoder-decoder model for skill acquisition and a conditional diffusion policy for action prediction. In the first stage, the skill encoder and decoder learn a codebook of latent skill vectors corresponding to languages conditioned on states. The diffusion policy then predicts the subsequent action directly, conditioned on the current state and latent skill generated by the skill encoder. An overview of our approach is depicted in Figure \ref{fig:overall}.

\subsection{Problem Formulation}
\label{sec:problem}

We consider learning in general environments modeled as the Markov decision processes (MDPs). In each environment, we are provided with an offline dataset consisting of $N$ demonstration sequences obtained from a diverse set of tasks using a behavior policy. Each trajectory consists of state-action pairs with one language label over $T$ time steps.

For multi-task environments, each language describes a set of tasks with varying quantities. The states and actions performed by the agent were stored as pairs along with a single language instruction in each trajectory . 
$$\tau_{i}=\left\{s_0,a_0,s_1,a_1,...,s_T,a_T,l\right\}_{i=0}^{N}.$$
\subsection{Mutual Information Skill Learning in LCSD}
\label{sec:MI}
In language-based imitation learning environments, the agent executes actions based on tasks specified through language. Therefore, the skills we learned must closely relate to the language instructions. $I(\cdot;\cdot)$ and $H(\cdot)$ refer to mutual information and Shannon entropy. Firstly, we directly maximize the mutual information between skills and language $I(\mathbf{z},l)$, where $\mathbf{z}$ represents skill sets for the entire trajectory. In multi-task environments, a single language may involve multiple skills, as shown in Figure \ref{fig:fisrt}. In such cases, skills need to segment the trajectory into sub-tasks based on different states. For example, when executing the instruction \textit{open the drawer and pick up the cup}, our skill needs to distinguish the current task of the agent based on whether the drawer is already open or not. To this end, we further aim to maximize the mutual information between skill and language conditioned on the current state, denoted as $I(l;z|s)$. In summary, our goal is to maximize:
\begin{align}
\label{eq:MI_first}
\mathcal{F}&=I(\mathbf{z};l)+I(l;z|s) \nonumber \\
&=H(l)-H(l|\mathbf{z})+H(z|s)-H(z|l,s) \\
\label{eq:MI_two}
&=H(z|s)+\mathbb{E}_{z \sim p, s \sim D}[\log p(z|s,l)] + \notag \\
&\mathbb{E}_{z \sim p, l \sim D}[\log p(l|\mathbf{z})]+\text{Const},
\end{align}


As shown in Equation \ref{eq:MI_first}, we use forward form on $I(\mathbf{z};l)$ and reverse form on $I(z|l,s)$. $H(l)$ represents the entropy of language instructions, which is constant in our offline dataset. The second term focuses on how our skills are related to language. The third term expects our skill distribution to have high entropy for better generalization conditioned on states. For the last term, our goal is to map deterministic discrete skills with current state and language instructions as conditions.


In Equation \ref{eq:MI_two}, we express the formula in the form of a probability distribution, where skills are sampled from a uniform distribution $p(z)$, and states $s$ and language $l$ are sampled from the stationary offline dataset. We implicitly optimize $H(z|s)$ by initializing unused codes for a broader range of skill selection and explicitly approximate the lower bound of conditional probability distribution by neural networks, the skill encoder $p_{\theta}(z|l,s)$ and 
the skill decoder $q_{\phi}(l|\mathbf{z})$. The encoder constrains the predicted skills for each step, while the decoder updates macroscopic language instruction reconstruction after skill generation on the entire trajectory. The ultimate optimization goal can be simplified as maximizing the lower bound of our objective $\mathcal{F}(\theta,\phi)$:
\begin{equation}
\label{eq:MI_last}
    \mathcal{F}(\theta,\phi) \geq H(z|s)+\mathbb{E}[\log p_{\theta}(z|s,l)]+\mathbb{E}[\log q_{\phi}(l|\mathbf{z})] .
\end{equation}

Our method aimed to enhance the diversity of skill codes by enabling the skill encoder to leverage a broader range of codebook latents. To achieve this, we explicitly augmented $H(z)$ from Equation \ref{eq:MI_last} using the proposed skill reinitialization approach mentioned below.

\subsection{Skill learning}
\label{sec:skill}

VQ-VAE is an unsupervised generative model for representation learning that uses an encoder to map images into latent space and a decoder to reconstruct the original image. In previous works on imitation learning, a skill encoder was used to directly map states to skills without a decoder, resulting in unstable, non-interpretable skills for task analysis \cite{skilldt, lisa}. \cite{choreographer} used the complete VQ-VAE framework for skill discovery, where a decoder was used to reconstruct states for computing rewards to update the world model in Actor-Critic training. LISA \cite{lisa} introduced language into VQ training to solve decision-making problems with IL. However, a single encoder mapping discrete skills from language-state embeddings is inadequate in learning the direct relation between skills and languages, resulting in poor stability in different environment settings(Figure \ref{fig:word_LORel}). More LISA skill maps are shown in the Appendix.

To address this, we jointly map the state and language to latent skill vectors when selecting skills, following VQ-VAE. The skill encoder $p_{\theta}(s,l)$ learns as a language-state representation. In vector quantization (VQ), a codebook comprising $M$ latent codes of skill vectors $z^{1...M}$ is utilized, and the skill vector closest to the encoder output is selected.

\textbf{Instruction Semantic Recovery:}
In language-based multi-task environments, we propose a language reconstruction-based VQ-VAE to better learn the specific tasks corresponding to skills in different states. We introduce a decoder corresponding to the last term of our optimization goal (Equation \ref{eq:MI_last}), which solely aims to align the skill vectors with the language representation. Unlike previous work \cite{choreographer}, our decoder does not participate in subsequent policy updates but rather serves to optimize the correspondence between the embeddings generated by the skill encoder and the language.

\begin{figure}[H]
	\centering
    \includegraphics[width=1\linewidth]{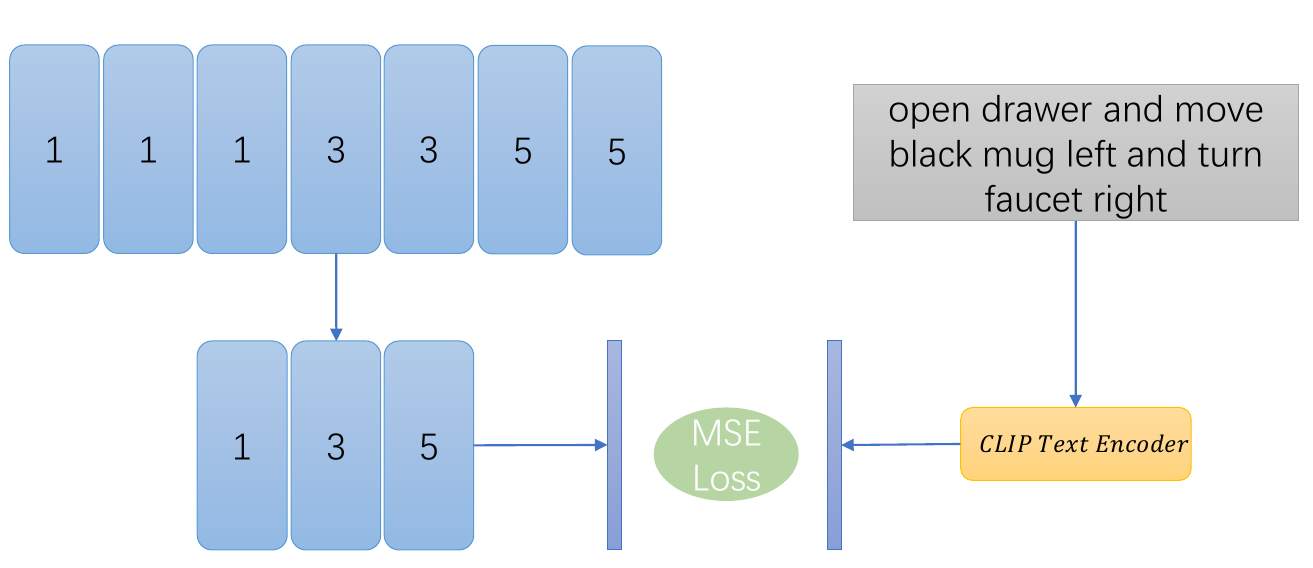}
    \caption{\textbf{Instruction Semantic Recovery Diagram}. The decoder's objective is to choose a distinct skill from each consecutive group within a trajectory and calculate the mean squared error (MSE) loss using the frozen CLIP \cite{clip} language embedding.}
    \label{fig:lang}
\end{figure}

We present the structure of our skill decoder in Figure \ref{fig:lang}. As language contains varying amounts and types of skills in different trajectories, we consider consecutive selections of the same skill as a sub-task and choose the first code from every consecutive group of equivalent skill codes. The decoder takes a discrete set of skills as input and outputs a vector, which is compared to the language vector obtained from a frozen CLIP text encoder \cite{clip} using MSE loss.
\begin{equation}
    \begin{split}
    L_{skill}&=-\log q_{\phi}(l|\mathbf{z}) - \log p_\theta(z|s,l)   \\
    &=\left\|q_{\phi}(U(z_q(s,l)))-E(l)\right\|_2^2 + \\
    &\beta\left\|p_{
    \theta
    }(s,l)-sg(z_q)\right\|_2^2 .\\
    \end{split}
    \label{eq:skill}
\end{equation}
 $U$ represents the unique selection of discrete skills, and $E$ refers to the CLIP text encoder that is frozen during training. The skill loss is comprised of two components: the reconstruction loss and the commitment loss, both of which are incorporated in the VQ training procedure according to Equation \ref{eq:VQ-VAE}. By minimizing the skill loss, we simultaneously maximize the Mutual Information based on Equation \ref{eq:MI_last}. 

\textbf{Skill Reinitialization:} Selecting a single code may lead to index collapse during the skill learning period due to preferential selection \cite{index_collapse}. We also encountered similar situations in some environments during evaluation, as shown in the upper image in Figure \ref{fig:word_LORel}. To address this issue, we used codebook reinitialization, which aims to involve more codebook vectors in skill learning and code selection. Inspired by \cite{choreographer}, We recorded times of codes selected within a certain training iteration and reinitialized codes with proper skill encoder vectors. Reinitialized codes can directly participate in skill selection and updates in subsequent training.

 We selected encoder outputs $p_\theta(s_i,l)$ to reinitialize inaccessible code $z_q^k$ with a probability of $d_p^2(z_q^k,s_i,l)$
 $\frac{d_p^2(z_q^k,s_i,l))}{\sum_s d_p^2(z_q^k,s_i,l)}$, Where $i$ is the index corresponding to the output of the encoder that selected the replacement code $p_{\theta}(s,l)$ is the output from the skill encoder on the last training batch, $N$ represent the total number of skills in the codebook, and $d_p$ is the fraction of the Euclidean distance between the encoder output and codebook embedding. Unlike \cite{choreographer}, we choose an embedding closer to the current code for more stable skill generalization.
\begin{equation}
     d_p^2(z_q^k,s_i,l)=\frac{1}{\left\|p_\theta(s_i,l)-z_q^k\right\|_2^2} .
\end{equation}

Unlike previous works, our approach incorporates a generalized strategy of not only initializing unused codes but also resetting the entire code set with a specified probability. This decision is motivated by our observation that inactive skills during the initial training phase can lead to significant instability in the early stages of training. To determine the reset probability, we calculate it proportionally to the ratio of each skill to the average skill selection. By adopting this method, we aim to enhance the stability and effectiveness of the training process.

\begin{equation}
z_q^k = \begin{cases}	
     p(\frac{d_p^2(p_\theta(s_i,l))}{\sum_s d_p^2(p_\theta(s,l))})*p_\theta(s_i,l), & y > \frac{M_k*N}{\sum_{j=1}^N M_j}, \\
    z_q^k, & y < \frac{M_k*N}{\sum_{j=1}^N M_j}

\end{cases}
\end{equation}

Where $M$ represents the number of times each skill is selected during this training session, and $y$ is a randomly generated float value ranging from 0 to 1. We aim to initialize the code with fewer prior selections, enabling more efficient updates in subsequent training iterations. We anticipate improved training efficiency and effectiveness by prioritizing initializing less frequently selected codes. Notably, our skill reinitialization method only takes place in the first 200 epochs of training, as we aim to make the most of each skill as possible while maintaining the way skills are learned. Therefore, fewer steps to initialize can avoid excessive human intervention in training and achieve better results.

\subsection{Diffusion policy for Imitation Learning}

We adopt the Denoising Diffusion Probabilistic Model (DDPM) \cite{DDPM} as our policy base model. The denoising network aims to predict the random noise added to the action in each iteration. The noisy input in each iteration can be formulated as $\mathbf{a}_i=\sqrt{\bar{\alpha}_i} \mathbf{a}+\sqrt{1-\bar{\alpha}_i} \boldsymbol{\epsilon}$, where $\bar{\alpha}$ are process variances, and random noise $\boldsymbol{\epsilon}$ is sampled from a Gaussian distribution $N(0,I)$.As an imitation policy to solve language condition tasks, our diffusion model can support language or skill information along with the current state as input simply by modifying the conditional input dimension. We modified the policy training loss as follows:
\begin{equation}
\begin{split}
     \mathcal{L}_{ddpm-s}(\theta)=\mathbb{E}_{\boldsymbol{\epsilon},i,s,a,l,\boldsymbol{z}}\left[\left\|\boldsymbol{\epsilon}-\boldsymbol{\epsilon}_\theta\left(\mathbf{a}_i, \boldsymbol{s},\boldsymbol{z}\right)\right\|^2\right] . \\
     \text{s.t.}\quad\boldsymbol{\epsilon},i \sim \mathcal{U}, ({s},{a},{l}) \sim \mathcal{D}, \boldsymbol{z} \sim z_q(s,l)
\end{split}
\label{eq:reinit}
\end{equation}
Where $i$ is sampled from $\mathcal{U}[1,T]$, denoise network $\epsilon_\theta$ is trained to predict random noise with state, action noise, and skill(or language) as input.

To combine skill (or language) and image features, we used different linear layers similar to the Temporal U-Net as our diffusion denoising network. For each MLP block, separated linear layers were used to unify the dimensions of the action noise, state, timestep, and skill embedding (language), and then they were added together. The final linear layer of the network outputs noise with the same dimension as the action. This network was designed to fully utilize conditional information. The detailed structure is shown in the Appendix.
\begin{algorithm}[tb]
\caption{LCSD training algorithm}
\label{alg:LCSD}
\textbf{Initialized Model}: diffusion policy $\pi$, skill encoder $p_\theta$, skill decoder $q_\phi$, CLIP encoder $\mathcal{E}$, Codebook quantize on encoder $\mathrm{q}$
\begin{algorithmic}[1] 
\WHILE{training iterations $i=1...N$}
\STATE \textit{-----Skill learning Period-----}
\STATE Sample batch $\tau=\{l,(s_0,a_0),...(s_T,a_T)\}_{i=0}^B$
\IF {Skill learning}
\FOR{each trajectory $\tau$}
\STATE $z_{0:T}\leftarrow p_\theta(s_{0:T},\mathcal{E}(l))$
\STATE record unselected codes in list $u$
\ENDFOR
\STATE Compute skill loss $\mathcal{L}_{skill}$ with Equation \ref{eq:skill}
\IF{$i<reinitupdate$ and $i$ mod $reinitstep=0$}
\STATE reinitialize unused code in list $u$ with probability on Equation \ref{eq:reinit}.
\ENDIF
\ENDIF
\STATE \textit{-----Behavior Cloning Period-----}
\IF{Skill learning}
\STATE $a^{'}_{0:T}=\pi(s_{0:T},\mathrm{q}(p_\theta(s_{0:T},\mathcal{E}(l))))$
\ELSE
\STATE $a^{'}_{0:T}\leftarrow \pi(s_{0:T},\mathcal{E}(l))$
\ENDIF
\STATE Compute Behavior cloning loss $\mathcal{L}_{ddpm-s}$
\STATE update with $\mathcal{L}_{LCSD}=\alpha\mathcal{L}_{skill}+\gamma\mathcal{L}_{ddpm-s}$
\ENDWHILE
\end{algorithmic}
\end{algorithm}

\begin{table*}
 
  \centering
  \begin{tabular}{lcccccc}
  \toprule
    Task  & Original & Lang+DT  & LISA  & LISA+init & Lang+Diffusion &  LCSD \\
    \hline
    BabyAI GoToSeq & 40.4 $\pm$ 1.2  &  62.1 $\pm$ 1.2 & 65.4 $\pm$ 1.6  & $\text{-}$ & 65.2 $\pm$ 8.6 & \textbf{67.8 $\pm$ 8.2}  \\   
    BabyAI SynthSeq &32.6 $\pm$ 2.5& 52.1 $\pm$ 0.5 & 53.3 $\pm$ 0.7  & $\text{-}$ & 55.1 $\pm$ 2.5 & \textbf{57.6 $\pm$ 2.2} \\
    BabyAI BossLevel & 28.9 $\pm$ 1.3 & 60.1 $\pm$ 5.5 & 58.0 $\pm$ 4.1  & $\text{-}$ & 55.0 $\pm$ 3.4 & \textbf{60.5 $\pm$ 7.4 } \\
    LORel sawyer state & 6 $\pm$ 1.2 & 33.3 $\pm$ 5.6 & 6.7 $\pm$ $3.3^*$  & 43.4 $\pm$ 0.2 & 43.0 $\pm$ 1.5 & \textbf{60.2 $\pm$ 5.7} \\
    LORel sawyer obs & 29.5 $\pm$ 0.07 & 15.0 $\pm$ 3.4 & 10.3 $\pm$ $1.4^*$  & 24.5 $\pm$ 4.3 & 36.6 $\pm$ 3.8 & \textbf{45.5 $\pm$ 5.1} \\
    CALVIN & 32.5 $\pm$ 2.5 & 11.7 $\pm$ 0.8 & 10.1 $\pm$ 3.3 & 10.9 $\pm$ 0.4 & \textbf{37.5 $\pm$ 2.6} & 35.6 $\pm$ 1.8 \\
    \hline
  \end{tabular}
   \caption{\textbf{Overall Performance}. We show our LCSD's evalution success rate (in \%) in different environments compared to the baseline imiation learning methods and skill learning method(LISA). The best method is shown in \textbf{bold}. We optimize LISA with official code from \cite{lisa} but cannot get normal performance on LORel due to index collapse, as indicated by * in our results.}
  \label{overall-table}

\end{table*}

LCSD is an end-to-end imitation policy, and we provided an overall structure feature in Figure \ref{fig:overall}. We developed a high-level skill generator based on a VQ-VAE model, which discretizes the latent space. The generated skills were then used in a diffusion policy as conditional information to predict the next-step action. The overall LCSD implimentation algorithm is shown Algorithm~\ref{alg:LCSD}. The overall loss combines skill and imitation policy as: 
\begin{equation}
\mathcal{L}_{LCSD}=\alpha\mathcal{L}_{skill}+\gamma\mathcal{L}_{ddpm-s}
\end{equation}
where $\alpha$ and $\gamma$ are used to balance the behavior cloning (BC) and skill learning losses.

\section{Experiments}


\subsection{Tasks}
\label{sec:task}
To verify the LCSD's effectiveness, we selected three benchmarks: BabyAI navigation \cite{babyai}, LORel Sawyer dataset \cite{LORel} and CALVIN robot tasks \cite{calvin}, which are all language-based and imitation learning environments without reward. Other benchmarks either lack language conditioning settings \cite{metaworld, kitchen} or focus on single-task environments with complex observation representations that generate hardly interpretable skills \cite{cliport}.

For the BabyAI benchmark, we used 10k trajectories evaluating three challenging tasks, namely GoToSeq, SynthSeq, and BossLevel. We collected an offline dataset of 50k trajectories on LORel and evaluated the performance on several task settings. 
For the CALVIN benchmark, we directly select 1216 trajectories from the CALVIN-D dataset relevant to the six tasks we modified. To eliminate interference on image encoders and focus solely on evaluating the underlying policy, we directly select the 21-dimensional perspective state of the CALVIN environment as observation input. More information on datasets is shown in the Appendix. 

\subsection{Baselines}

We compared our proposed LCSD with several baselines in our experiments:

\textbf{Original}: The BC baselines from original papers on three benchmarks. In BabyAI we
adopt their RNN-based method \cite{babyai}. In the LORel environment, we compared with the planner algorithm as language conditioned BC baseline. We trained MULC from \cite{calvin} on our CALVIN setting by changing the vision encoder into a simple MLP for perspective state observation.

\textbf{Language conditioned DT policy}: A behavior cloning Decision Transformer (DT) \cite{DT} based policy that takes the language instruction and past observations as inputs to predict action.

\textbf{LISA} \cite{lisa}: A hierarchical imitation learning structure based on a skill encoder and DT based policy.

\textbf{LISA with code reinitialization}: LISA with code reinitialize to better generalize skill code, denoted as LISA init. 

\textbf{LCSD on DT policy}: We apply the skill learning method of LCSD to the DT policy, referred as  LCSD+DT.

\textbf{Language Condition Diffusion Policy}: An imitation learning structure with diffusion policy condition on language. Different from LCSD, the input of the diffusion model is language tokens generated by the CLIP text encoder and current observation. We modify the structure by directly minimizing the behavior cloning loss in Equation \ref{ddpm-lang}. 
\begin{equation}
\label{ddpm-lang}
    \mathcal{L}_{ddpm-l}(\theta)=\mathbb{E}_{\boldsymbol{\epsilon},i \sim \mathcal{U},(\boldsymbol{s}, \boldsymbol{a},\boldsymbol{l}) \sim \mathcal{D}}\left[\left\|\boldsymbol{\epsilon}-\boldsymbol{\epsilon}_\theta\left(\mathbf{a}_i, \boldsymbol{s},\boldsymbol{l}\right)\right\|^2\right] .
\end{equation}

\subsection{Results}

We evaluated our approach in three environments. BabyAI serves as the most straightforward task with discrete actions for 2D navigation, while LORel is a medium-difficulty multi-task language environment based on Metaworld. 
Table \ref{overall-table} presents the overall results for different tasks. All baseline algorithms were trained over 1500 iterations using three distinct seeds to ensure the reproducibility and diversity of results. For a more statistically significant analysis, our LCSD was tested with five different seeds. Notably, LCSD outperformed the other language condition BC methods in various tasks. Specifically, LCSD showed superior performance in multi-task and complex LORel environments.


\begin{table*}
 
  \centering
  \begin{tabular}{lc|ccc|ccc}
  \toprule
    Language Settings & Observations & LISA & LISA+init & LCSD+DT  & Lang+Diffusion & LISA+Diffusion & LCSD \\
    \hline
    
    \multirow{2}{*}{Seen} & State & 6.0 &  43.4 & 43.3  & 46.6 & 30.0 & \textbf{60.2}  \\   
     & Image & 3.5 & 23.5 & 30.3 & 36.5 & 31.2 & \textbf{50.8}  \\ 
     \hline
     \multirow{2}{*}{Unseen verb} & State & 0.0  & 31.5 & 20.0 & 26.5 & 33.5 & \textbf{50.2}  \\   
     & Image & 10.0  &  14.1 & \textbf{30.6} & 23.1 & 29.5 & 30.4  \\ 
     \hline
     \multirow{2}{*}{Unseen noun} & State & 4.7  & 27.5 & 24.1 & 24.5 & 20.0 & \textbf{29.3}  \\   
     & Image & 13.5 &  20.0 & 24.4 & 26.8 & 20.5 & \textbf{27.5}   \\ 
     \hline
     \multirow{2}{*}{Unseen verb \& noun} & State & 3.5 & 24.5 & 26.6 & 17.5 & 27.0 & \textbf{28.7}  \\   
     & Image & 7.2  &  27.4 & 33.3 & 13.7 & 27.9 & \textbf{36.5}   \\ 
     \hline
      \multirow{2}{*}{Human} & State & 1.5  &  23.5 & 25.0 & 26.2 & 25.0 & \textbf{35.5}   \\   
     & Image & 7.5  & 24.5 & 32.5 & 23.7 & 25.2 & \textbf{33.4} \\ 
     \hline
     \multirow{2}{*}{Overall} & State & 3.2  & 30.1 & 27.8 & 28.3 & 27.1 & \textbf{40.8}   \\   
     & Image & 8.3  &  22.0 & 30.2 & 24.8 & 26.9 & \textbf{35.8} \\ 
    \hline
  \end{tabular}
   \caption{\textbf{The Performance in LORel Sawyer Environment on different task settings}. The evaluation succcess rate (in \%) of different algorithms on different task settings on LORel Sawyer state and image environments. The three algorithms on the left are all based on the DT model, while the three on the right are based on the diffusion model.}
  \label{tab:LORel-table}

\end{table*}

\begin{table}[H]

\centering
  \begin{tabular}{lrrrr}
  \toprule
    \multirow{2}{*}{CALVIN Tasks} &\multicolumn{2}{c}{DT model}  & \multicolumn{2}{c}{Diffusion model} \\
    &language  & skill & language & skill \\
        \hline
    Turn on ledbulb & 0 & 0 & 0.63 & 0.13 \\
    Turn off ledbulb & 0.25 & 0.13 & 0.25 & 0.25 \\
    Move slider left & 0 & 0.13 & 0.12 & 0.38 \\
    Move slider right & 0 & 0 & 0.5 & 1.0 \\
    Open drawer & 0 & 0 & 0.25 & 0.13 \\
    Close drawer & 0.25 & 0.25 & 0.5 & 0.25 \\
    \hline
    Overall & 0.083 & 0.085 & \textbf{0.375} & \underline{0.357} \\
    \hline
\end{tabular}
\caption{Success rate on CALVIN tasks}
\label{table:calvin}
\end{table}

\textbf{Diffusion model can leverage stability in difficult tasks:} While serving as a long-horizon benchmark, the language label in  CALVIN corresponds to a single skill, which is different from the other two benchmarks (See appendix for more dataset details). We mainly introduce CALVIN to evaluate the performance of different imitation learning models on difficult tasks rather than to measure the effect of skill learning. The dataset we selected for CALVIN only contains only 1216 trajectories, making the tasks even more challenging.


Our diffusion policy performed well in different tasks without requiring special modifications to the parameters, particularly excelling four times in CALVIN tasks (Comparing DT and Diffusion column in Table \ref{overall-table}).

Table \ref{table:calvin} lists the success rates in six different tasks, clearly showing that the diffusion-based policy outperforms DT-based models. It is typical for language-based models to exhibit slightly superior performance compared to skill-based models as a result of employing limited training data in CALVIN, along with massive redundant data. More detailed information about CALVIN's task settings can be found in the Appendix.



\begin{figure}
            \centering
		\includegraphics[width=1\linewidth]{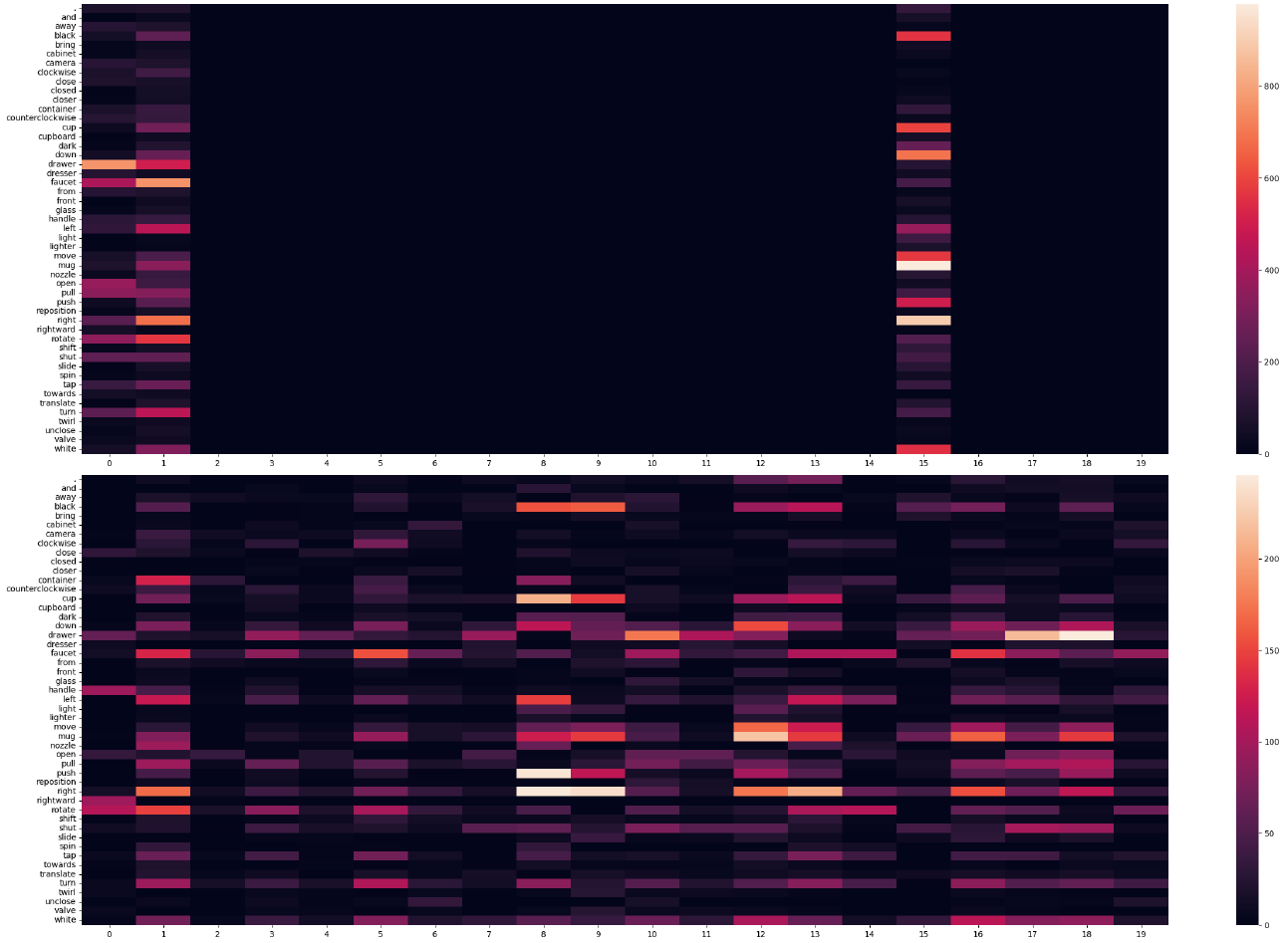}
		\caption{Skill-language mapping in LORel state environment. Up: skill-language graph on LISA (single encoder); Down: skill-language diagram of our LCSD.
            }
		\label{fig:word_LORel}
\end{figure}

\textbf{Skill Visualization:} To demonstrate the specific meaning of the discrete skills generated by our algorithm, we recorded the correlation between language and the selection of skill codebook during evaluation, as done in \cite{lisa}. We first show the skill map of \cite{lisa} in the upper image and find that most of the skills in the codebook are not involved in training. Different tasks can only be divided into two types of skills, which cannot be effectively trained, lea-ding to index collapse. The lower image in Figure \ref{fig:word_LORel} shows that all 20 codes were selected during evaluation, with a strong correspondence observed between language tokens and skill codes. For instance, Skill code 19 (the nineteenth column) corresponds to the action "\textit{turn/rotate faucet right/clockwise}," while skill code 0 (the zero column) represents "\textit{rotate handle rightward and open drawer}." It is reasonable for a single skill to indicate these two tasks because in LORel, the faucet is placed before the drawer, making it convenient for the agent to move the handle to the right while opening the drawer. More detailed skill maps are shown in the Appendix.

\subsection{Ablation Study}

\begin{figure}
	\centering
    \includegraphics[width=1\linewidth]{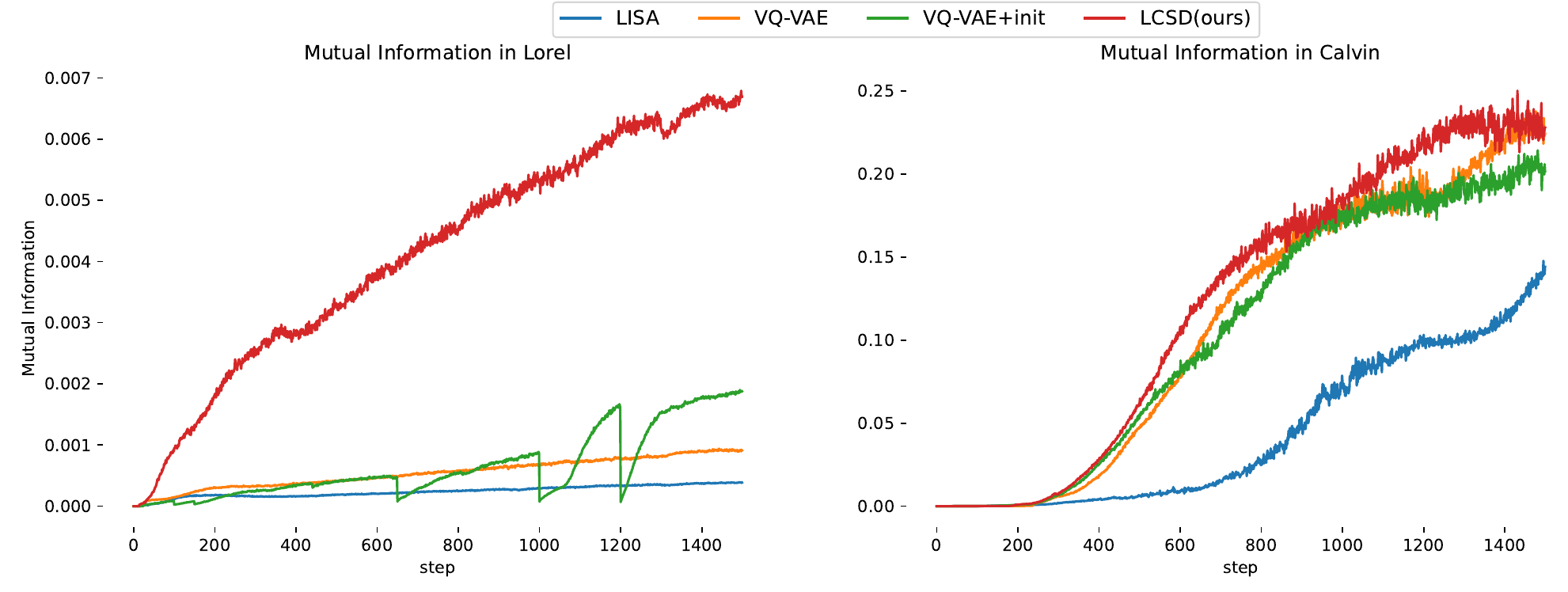}
    \caption{\textbf{MI training curve in CALVIN and LORel with difference skill learning methods}. We show the mutual information curves of our method during training in different environments on different skill learning methods.}
    \label{fig:MI}
\end{figure}

\textbf{Importance of Code Reinitialization and Instruction Recovery:} To assess the effectiveness of our skill-learning method, we conducted an analysis by plotting mutual information (MI) curves for different models to examine the correlation between language and skills. Figure \ref{fig:MI} illustrates the MI curves for four skill acquisition methods. Notably, the utilization of a skill-language decoder and skill reinitialization leads to a more pronounced increase in mutual information. It is important to highlight that, in the LORel dataset, each trajectory corresponds to multiple sub-tasks, necessitating a stronger correlation between language and skills when compared to the CALVIN environment. To clarify the specific meaning of discrete skills, we plot corresponding word frequency on LCSD with and without code reinitialization and reconstruction in Figure \ref{fig:word_LORel}. The comparison between the two figures clearly indicates that without the support of these two methods, the selection of code skills is limited to a small number, which is also observed in the DT \cite{DT} model, with even greater severity. The usage of code reinitialization provided a significant improvement in this case. In Appendix we display more skill maps with different LCSD settings in different environments.

\textbf{Generality of skill learning method:} Our skill-learning method can be extended to different models. In DT-based models, we observed index collapse in LORel environments, leading to poor test results. However, this problem was resolved by adding code reinitialization (Table \ref{overall-table}). Table \ref{tab:LORel-table} shows more detailed experiment results on LORel. By comparing our approach with LCSD combined with DT, it becomes evident that our approach can be used for the DT-based model for better skill discovery. More skill frequency figures and results are shown in the Appendix.

\textbf{Stability in multi task settings and varying parameters:} We conducted detailed experiments on different types of settings in the LORel environment, as shown in Table \ref{tab:LORel-table}, to investigate whether language settings affect the model's performance in multi-tasks. By manipulating various words within sentences, we aimed to enhance skillful semantic comprehension. "seen tasks" refers to language descriptions that were identical to the training set. At the same time, "human" indicates completely different sentences that convey the same meaning. Our LCSD with code reinitialization and language recovery outperformed other methods in almost all the task settings. In VQ-VAE, the number of skill vectors in the codebook $M$ and the number of combined skill vectors in the language decoder are relatively essential parameters. However, we found our LCSD to be robust enough to these choices, as shown in the Appendix.

\begin{table}[H]
  
  \centering
  \begin{tabular}{lcccc}
  \toprule
    Policy & Timestep $N$  & CALVIN  & LORel & BabyAI  \\
    \midrule
    \multirow{4}{*}{DDPM} &25 & 297 & 623 & 560 \\   
    & 50 & 619 &	720 & 840 \\
    & 75 & 880 & 1023 & 1240  \\
    & 100 & 1200 & 1400 & 2000 \\
    \hline
    DDIM & - & 256 & 525 & 450 \\
    \hline
    DT & - & 225 & 801 & 500 \\
    \hline
  \end{tabular}
  \caption{\textbf{Inference time of LCSD and DT based model in three benchmarks} (second per episode). }
\label{inference-table}
\end{table}

\textbf{Inference time of Diffusion policy:}
The diffusion policy's inference phase is time-consuming and depends on the hyperparameter timestep $n$. Therefore, we explored the impact of different timestep values in the BabyAI environment in the appendix. When set to 100 in the experiment, the evaluation time is approximately two to three times longer than the DT-based model. We list different inference times on different timestep among three benchmarks in Table \ref{inference-table}. To ensure both efficiency and accuracy, we recommend that the n timestep should be defined as 50 for the experiment. For further study, we adopt the DDIM evaluation phase on our model and set the time step to 10. In this case, LCSD performed faster than DT during the evaluation phase and performed better than DDPM with low timesteps.

\section{Conclusion}
In this paper, we have presented LCSD, a novel skill-based imitation learning framework designed for the purpose of multi-task skill discovery and behavior cloning in a language-conditioned environment. Our approach has demonstrated excellent performance in generating discrete skills while aligning with language in different environments. By initializing with diverse codes and establishing a stronger connection between skills and language through the language decoder, we have achieved more accurate and stable skill representations.

\textbf{Limitations and Future Work:} Our approach does not analyze the interconnections between different skills, which can be crucial in multi-tasking problems that are typically decomposed into a series of related sub-tasks. It is an interesting avenue for future research, with the potential to learn powerful skills to extend to more unknown tasks.



\bibliography{aaai24}

\newpage
\section{Appendix}

\subsection{Algorithm Optimization}

\subsubsection{Practical Optimization:}

In Equation 3 of the main paper, we adopt $H(z)$ instead of $H(z|s)$ to optimize the latent skill representation. Skills are mostly irrelevant to the state of the instruction following 
 tasks.
Formally, we build the mutual-information between latent skill $z$ and $s$: 
$$I(z;s)=H(z|s)-H(z),$$ 

where $H(z|s)$ is the conditional entropy of the skill $z$ given the state $s$, and $H(z)$ is the entropy of the skill $z$. We assume that the mutual information between skill and state is low, in which case $H(z|s)$ can be approximated by $H(z)$. The entropy of skills measures the degree of randomness or disorder. As we introduce code reinitialization to make discrete skill selection rather than training only a few codes (as shown in Figure \ref{fig:lorel_lisa}), we implicitly increase the randomness of skills.

\subsubsection{Mutual Information entropy loss:}

Here, we explain how to transform the objective from maximizing entropy to minimizing MSE error and incorporate it into VQVAE training; we provide the following proof:
First, we have our optimal goal from Equation 3.

$\mathcal{F}(\theta,\phi)=H(z|s)+\mathbb{E}[\log p(z|s,l)]+\mathbb{E}[\log q(l|\mathbf{z})]$

In our implimentation we have skill encoder $p_{\theta}(z|s,l)$ and language decoder $q_{\phi}(\mathbf{z})$. The first term is optimized by code reinitialization, which is mentioned in the appendix. Here, to prove how we transfer maximizing conditional entropy into minimizing MSE loss. The conversion mode of the third term of the formula is illustrated below.

\[
\begin{aligned}
& \mathbf{E}[\log p(z \mid l)]=\sum_{i=1}^m \log p\left(l_i \mid z_i ; \theta\right) \\
& =\sum_{i=1}^m\left(\log \frac{1}{\sqrt{2 \pi} \sigma} \exp \left(\frac{\left(l_i-\mathrm{q}_\phi\left(z_i\right)\right)^2}{2 \sigma^2}\right)\right) \\
&=\frac{1}{2 \sigma^2} \sum\left(l_i-\mathrm{q}_\phi(z_i)\right)^2 -\mathrm{m} \ln \sigma \sqrt{2 \pi} \\
\end{aligned}
\]

By removing the constant term, we can equivalently maximize entropy, equivalent to minimizing the mean squared error (MSE) loss. In Equation 5, we optimize this term in each episode by concatenating skill into trajectory instruction and compute MSE loss with language embedding from the CLIP encoder. The second term can also be optimized with $|p_{\theta}(s,l)-sg(z_q)|_2^2$.

\subsection{Implementation Details}

We provide details on the implementation of our LCSD algorithm in this section.

\textbf{Language Encoder:} We use a pre-trained CLIP ResNet50 text encoder network as our language encoder. This part is used only to decouple the complex language and will not be involved in the subsequent updates.

\textbf{Observation Encoder:} We use linear MLP layers as our state encoder. In the LORel environment, image observations are encoded by a convolution network similar to the original paper.

\textbf{Skill Encoder:} We use linear MLP as our skill encoder with language embeddings and states as input. The output skills correspond to the states of each step.

\textbf{Skill Decoder:} We use a three-linear-layer MLP as our language decoder with unique skill sets as our input. The output vector is consistent with the dimensionality of the language embedding.

\begin{figure*}
  \centering
  \includegraphics[width=13cm]
{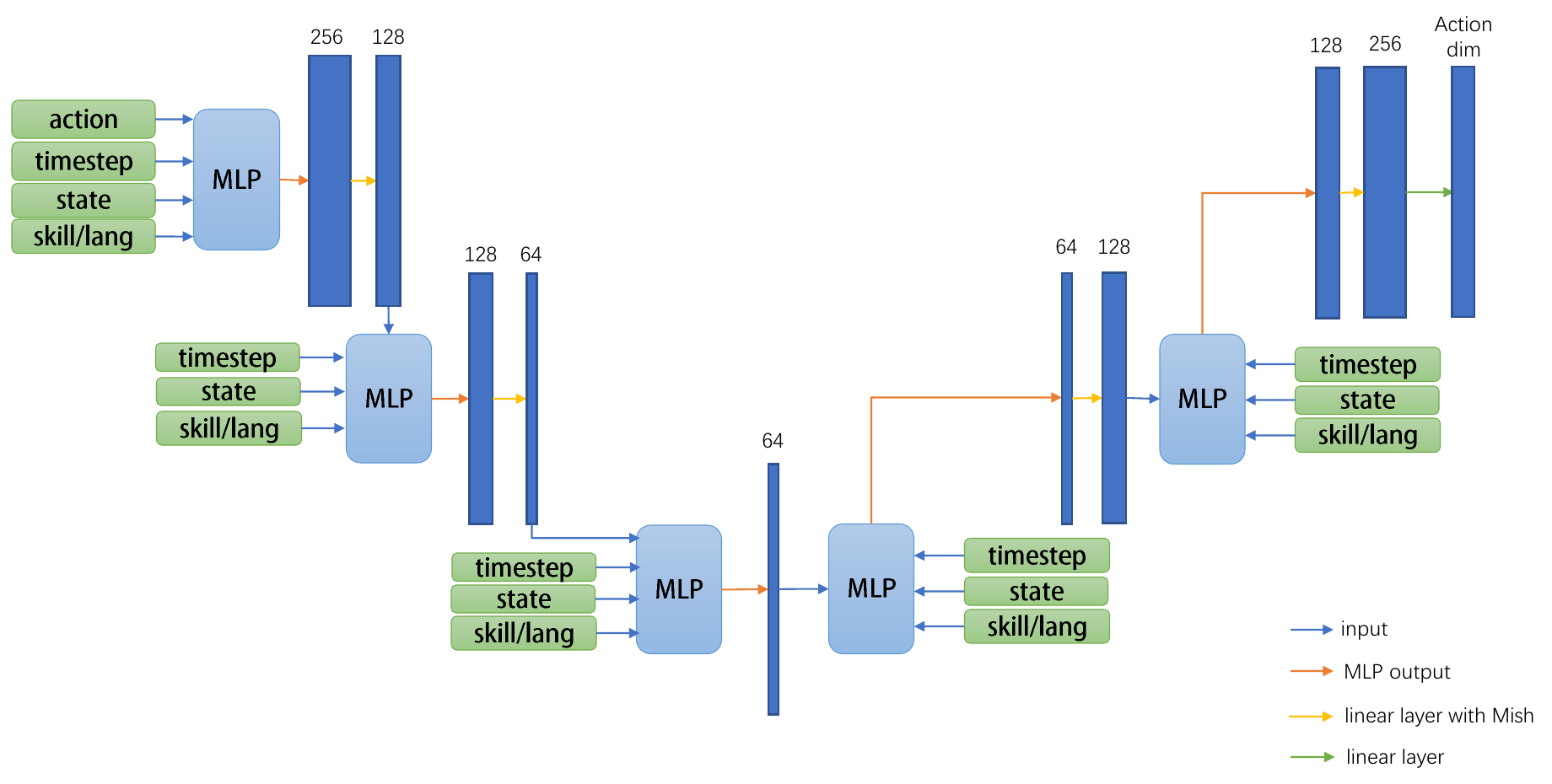}
  \caption{\textbf{Denoising Network Architecture}. We use a network similar to U-net. Different linear layers reduce and expand the spatial dimensions while fusing multimodal features. }
  \label{fig:network}
\end{figure*}

\textbf{Diffusion Model:} We define a conditional U-net model for fusing different modal information. This denoising model also extends to the language condition, which only requires changing the initialization dimension. The denoising network structure is shown in Figure \ref{fig:network}.

Our implementation uses the PyTorch framework on an NVIDIA A100 GPU. Parameters of that LCSD method used on different environments are listed in Table \ref{tab:Hyparameters}. For tasks of different difficulty, we use different sizes of codebook to measure. We use 20 skill codes for Babyai and Calvin as these environments do not contain many skills. For the LORel environment, we use 30 skill codes to represent more skills. Our method remains robust in varying hyperparameters, which is shown in Section \ref{app:ski}.

\begin{table}
\caption{\textbf{Parameters of LCSD}}
    \centering
    \begin{tabular}{l|ccc}
    \toprule 
    Hyperparameter  & BabyAI & LORel & Calvin \\
    \hline
         Behavior loss Weight & 5.0 & 5.0 & 5.0 \\
         Skill loss Weight & 2.0 & 2.0 & 5.0 \\
         Reconstruct loss Weight & 0.01 & 0.01 & 0.1  \\
         commitment Weight & 1.0 & 1.0 & 1.0 \\
         Batch size & 256 & 256 & 128 \\
         Codebook Dim & 16 & 16 & 16 \\
         Skill Number & 20 & 30 & 20 \\
         Reconstruct Option & 3 & 4 & 2 \\
         Policy lr & 2e-4 & 2e-4 & 2e-4 \\
         Skill lr & 1e-4 & 1e-4 & 1e-4 \\
         Timesteps & 50 & 100 & 50 \\
    \bottomrule
    \end{tabular}
    \label{tab:Hyparameters}
\end{table}

\subsection{Experimental Details}
We describe detailed information about the evaluation environments and the specific demonstration trajectories.

\begin{figure*}
  \centering
  \includegraphics[width=0.9\textwidth]{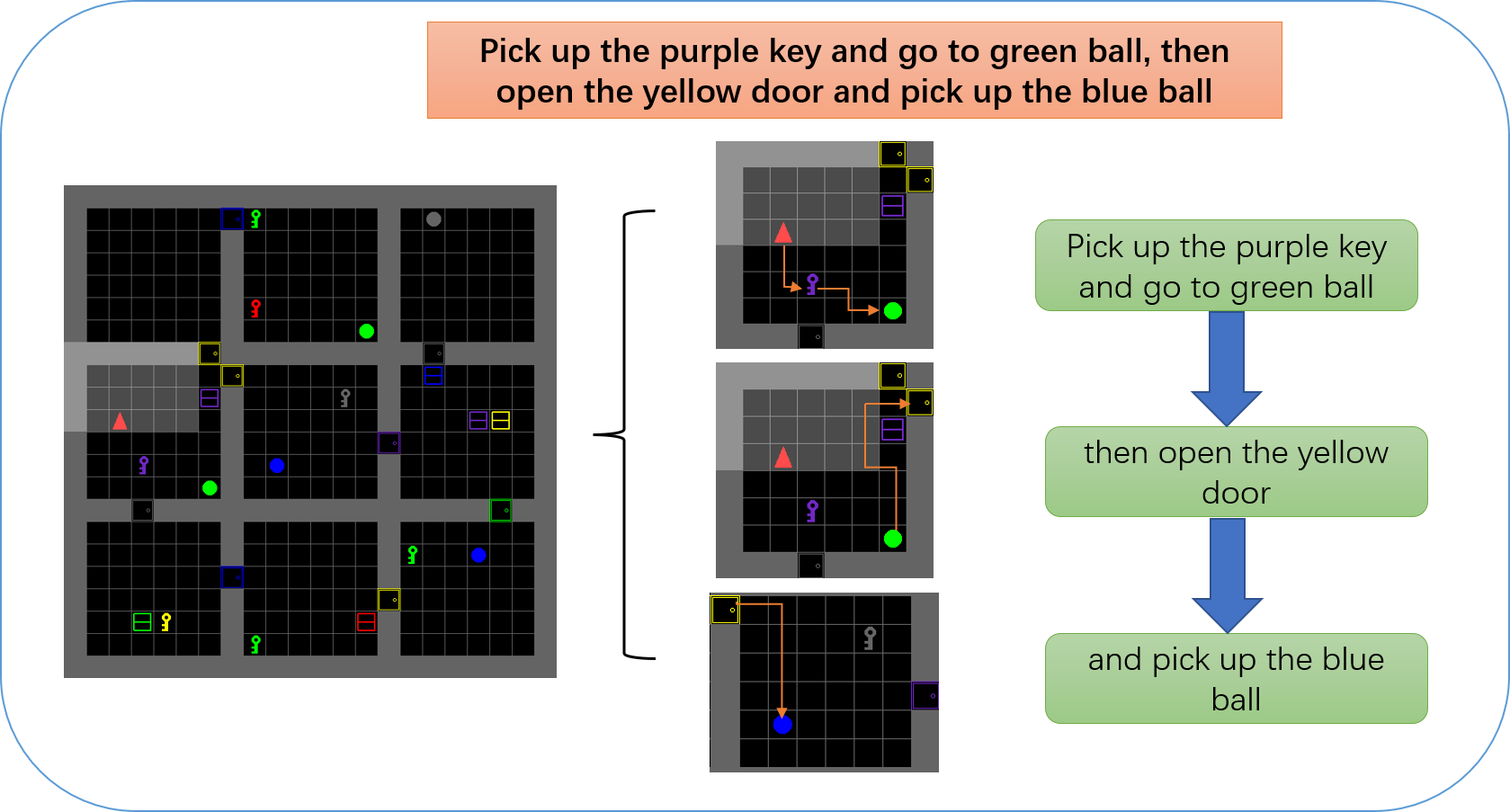}
  \caption{Babyai Bosslevel on task \textit{Pick up the purple key and go to green ball, then open the yellow door and pick up the blue ball }.}
  \label{fig:env_babyai}
\end{figure*}

\label{app:exp}
\textbf{Babyai}: Babyai is a two-dimensional grid game with multiple task settings, where the agent performs a series of subtasks according to the language instructions, such as "open the yellow door" and "go to the key behind you". To evaluate our skill-based model, we select three relatively challenging tasks:  GoToSeq, SynthSeq, and BossLevel. We use 10,000 trajectories for each task in our experiments. An example of BabyAI BossLevel sequence is shown in Figure \ref{fig:env_babyai}. 

\begin{figure*}
\centering
\includegraphics[width=0.9\linewidth]{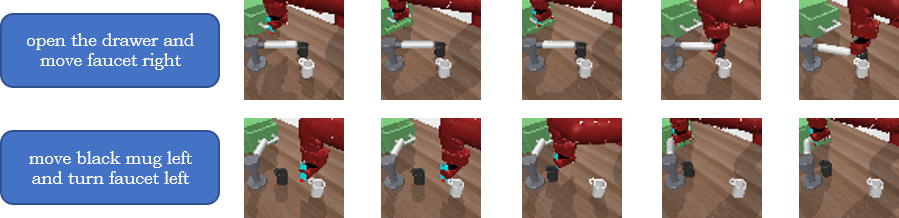}
\caption{LORel trajectory on two different tasks}
\label{fig:env_lorel}
\vspace{-5mm}
\end{figure*}

\textbf{LORel}: LORel is a robot manipulation environment based on Metaworld in a tabletop scenario, where a simulated Sawyer robot manipulates a faucet, a drawer, and two cups in different colors. Each trajectory is labeled with a language and contains one or more tasks, such as "Move black mug left and turn faucet left". Two trajectories of different tasks are shown in Figure \ref{fig:env_lorel}. We collect an offline dataset of 50,000 trajectories in our experiments from the official code.

\begin{table*}
  \caption{Calvin dataset language settings}
  \label{table:calvinlanguage}
  \centering
  \resizebox{\textwidth}{!}{
  \begin{tabular}{lllll}
    \toprule
    Catagory  & Detail tasks & Training Language  & Evaluation language & Num \\
    \midrule
    \multirow{3} * {lightbulb} & \multirow{3} * {turn on/off lightbulb} & toggle the light switch to turn on/off the light bulb & \multirow{3} * {use the switch to turn on/off the light bulb} & \multirow{3} * {95} \\
    ~ & ~ & turn on/off the light bulb &  ~ & ~ \\
    ~ & ~ & move the light switch to turn on/off the light bulb &  ~ & ~ \\
    \midrule
    \multirow{3} * {drawer} & \multirow{3} * {open/close drawer} & open/close the cabinet drawer & \multirow{3} * {pull/push the handle to open/close the drawer} & \multirow{3} * {303} \\
    ~ & ~ & grasp the drawer handle and open/close it &  ~ & ~ \\
    ~ & ~ & pull/push the handle of the drawer &  ~ & ~ \\
    \midrule
    \multirow{3} * {slider} & \multirow{3} * {move slider left/right} & grasp the door handle, then slide the door to the left/right & \multirow{3} * {push the sliding door to the left/right side} & \multirow{3} * {382} \\
    ~ & ~ & move the door all the way to the left/right &  ~ & ~ \\
    ~ & ~ & slide/push the door to the left/right &  ~ & ~ \\
    \midrule
    \multirow{3} * {light} & \multirow{3} * {push down the button to turn on/off the led} & press the button to turn on/off the led light & \multirow{3} * {push the sliding door to the left/right side} & \multirow{3} * {382} \\
    ~ & ~ & turn on/off the led light &  ~ & ~ \\
    ~ & ~ & toggle the button to turn on/off the led light &  ~ & ~ \\
    \midrule
    Unknown tasks & \multicolumn{3}{c}{Turn on/off green/yellow lamp  move/toggle the light switch to turn off the yellow/green light} & 227 \\
    \bottomrule
  \end{tabular}}
\end{table*}

\textbf{Calvin}: Calvin is a manipulation environment that uses a Franka Panda arm in four different settings. We adopt the tasks proposed by the original paper and modify the evaluation on six tasks: Open Drawer, Close Drawer, Turn on Lightbulb, Turn off Lightbulb, Move Slider Left, and Move Slider Right. We directly select 1,216 trajectories from the Calvin-D dataset that are relevant to the above six tasks as our offline dataset.
Since RGB image inputs depend on factors such as image encoders and multiple perspectives, in order to eliminate interference and focus solely on evaluating the underlying policy and their performance on DT, we directly selected the 21-dimensional perspective state of the Calvin environment as the input for LCSD. As this type of state space does not include the spatial aspect of the ball, we excluded tasks related to the ball in our task selection, which is mentioned above. The dataset can be downloaded from the official link.

\begin{figure}
  \centering
  \includegraphics[width=1\linewidth]{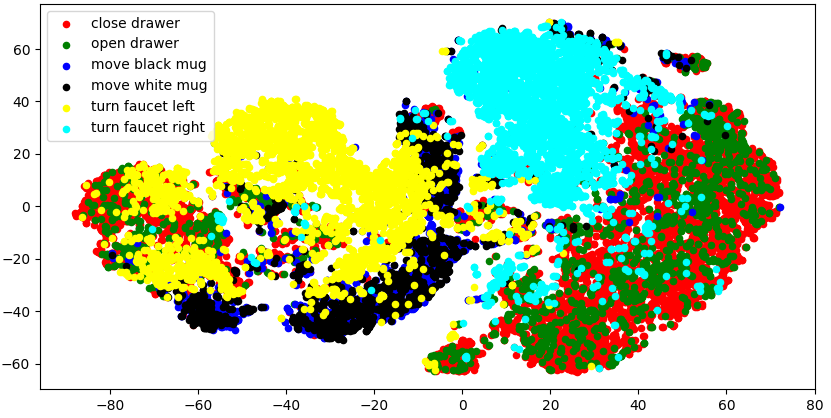}
  \vspace{-2mm}
  \caption{Data visualize in LORel.}
  \label{fig:lorel_visual}
\end{figure}

Table \ref{table:calvinlanguage} lists the language labels used in our training dataset for the Calvin environment. We observe that the language labels for each task in Calvin are not standardized but vary as much as possible, similar to the different test settings in LORel. However, unlike LORel, the training dataset in Calvin consists of diverse language instructions, making it more challenging to train our LCSD algorithm.

As a result, we find that the language condition model outperformed the skill condition in the Calvin environment, where we use a diverse offline dataset with varying instruction labels. This outcome is reasonable, given the nature of the training dataset in Calvin. It demonstrates the importance of considering the language condition in our LCSD algorithm when dealing with diverse and varied language instructions. We are pleased to see that our LCSD maintains a high level of performance in all tasks.

\subsection{Additional Experiments}

\subsubsection{Effect of Skill and Reconstruction Option Numbers:}

Codebook size and the number of reconstruction options are important hyperparameters that may affect skill learning performance. In our approach, the upper limit of the skill set used by the decoder needs to be adjusted depending on the task, similar to how codebook size is adjusted in VQ-VAE. In Figure \ref{fig:lcsdmulti}, we have shown the stability of skills generated by LCSD in varying codebook sizes. We further show our success rate in LORel environments in Table \ref{tab:lorel_parameter} to see that the overall result is stable in the process of parameter change. (Note that this success rate is an overall result of all tasks, which is different from the main paper.)

\begin{table}[H]
\caption{LCSD performance on LORel with different sizes of codebook and number of reconstruction options}
    \centering
    \begin{tabular}{c|ccc}
        \toprule 
    \diagbox{Codebook}{Option} & 3 & 4 & 5 \\
        \midrule
         20 & 35.10$\%$ & 35.34$\%$ & 38.92$\%$ \\
         30 & 37.40$\%$ & 38.70$\%$ & 36.36$\%$ \\
         40 & 37.66$\%$ & 38.96$\%$ & 37.92$\%$ \\
         \bottomrule
    \end{tabular}
    \label{tab:lorel_parameter}
\end{table}

In Table \ref{tab:lorel_parameter}, We find that the best result is achieved when LCSD with the setting of 40 skills and four reconstruction options. Meanwhile, we observe that the performance of each parameter fluctuated within a normal range. These results confirm the robustness of our LCSD algorithm on these two hyperparameters. It demonstrates that our algorithm can handle varying codebook sizes and the number of reconstruction options while still achieving stable performance.

\subsubsection{How do Diffusion timestep affect results:}
\begin{table}
\caption{Diffusion language policy performance on Babyai}
    \centering
    \resizebox{0.8\linewidth}{!}{
    \begin{tabular}{c|ccc}
        \toprule 
     Timestep & SynthSeq & GoToSeq & BossLevel \\
        \midrule
         100 & 54.1$\%$ & 65.2$\%$ & 55.2$\%$ \\
         50 & 52.2$\%$ & 65.4$\%$ & 48.5$\%$ \\
         \bottomrule
    \end{tabular}}
    \label{tab:Hyparameters}
\end{table}
Additionally, we evaluate the effect of timestep on the results of behavioral cloning in the DDPM process. A larger timestep can provide a more accurate distribution for the input recovery, but it will also take more time during evaluation. We evaluate our Diffusion language-conditioned policy on the BabyAI environment with two timestep settings. In the BabyAI environment, the performance of LCSD is not significantly influenced by timestep, as shown in Table \ref{tab:Hyparameters}.

\begin{figure}
  \centering
  \includegraphics[width=1\linewidth]{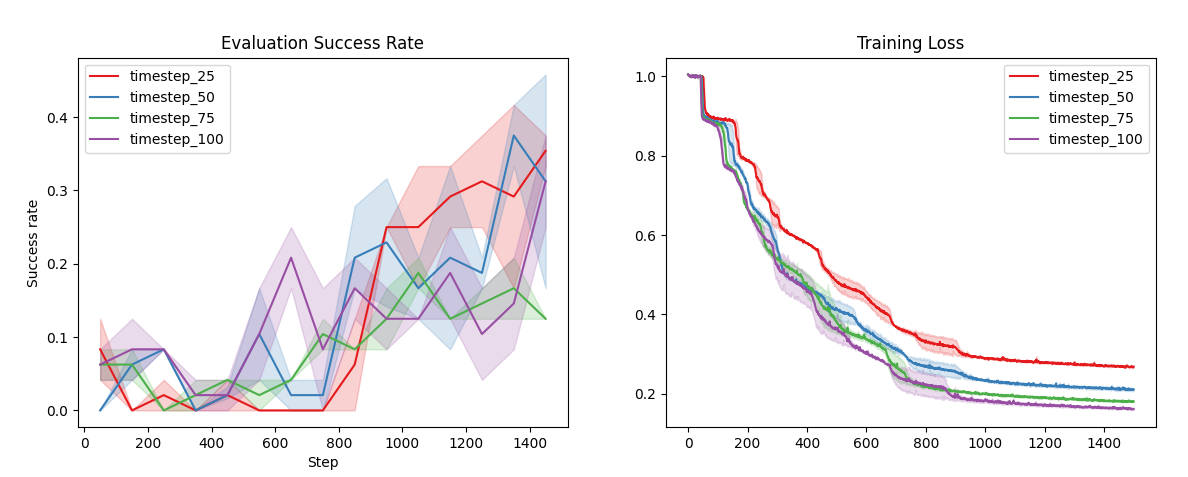}
  \vspace{-5mm}
  \caption{Diffusion performance and training loss on Calvin with different timesteps.}
  \label{fig:calvin_codes}
\end{figure}

We conduct more ablation experiments on the Calvin environment with 25, 50, 75, and 100 timesteps for six tasks, as shown in Figure \ref{fig:calvin_codes}. In the complex Calvin dataset, we observe that a larger timestep does not correspond to better results. As seen in the table, timesteps of 25 and 50 are the most appropriate in both performance and efficiency, while timesteps of 75 perform the worst, even though the training loss decreased more quickly. We believe this is because a good fitting is not always reflected in the performance in non-optimal diverse datasets. However, with our diffusion model, we can handle both large and precise as well as small and diverse datasets with appropriate timesteps.

\section{Skill Analysis}
\label{app:ski}

\begin{figure}
	\centering
    \subfigure[skill-word correlation map]{
        \includegraphics[width=0.99\linewidth]{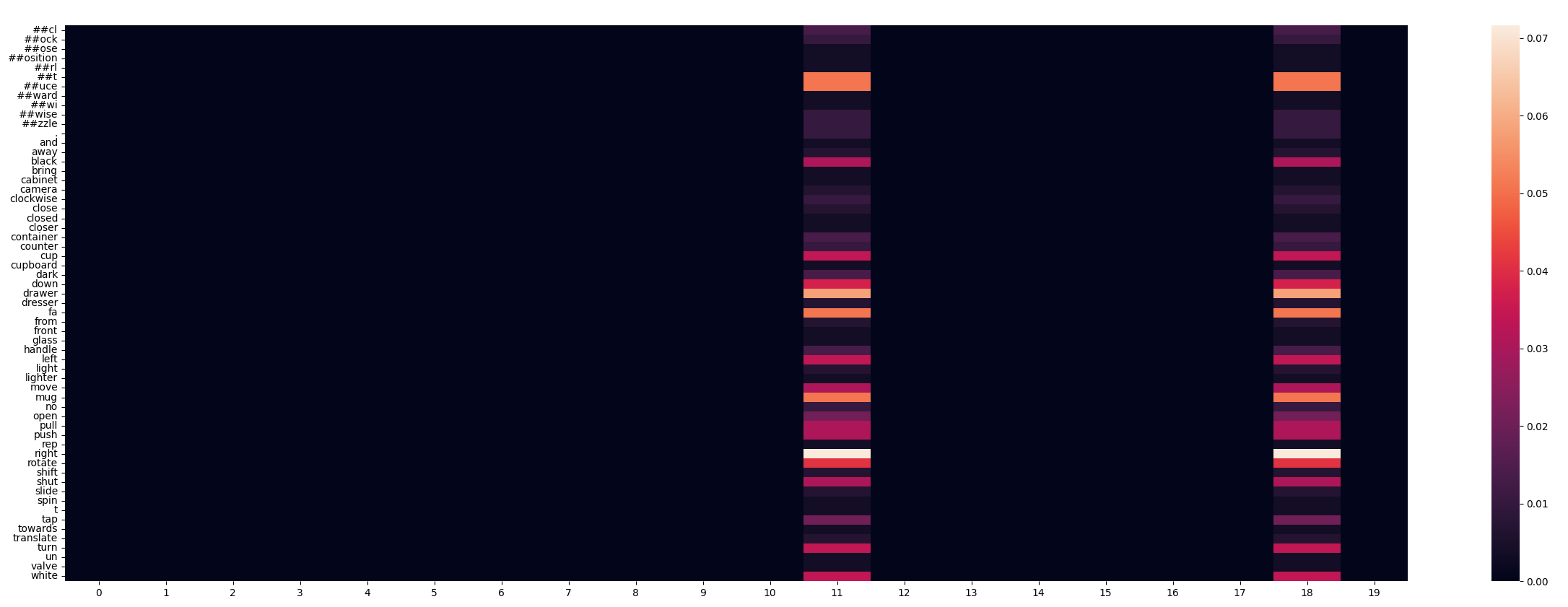}
        \label{fig:lorel_lisa}
    }
    
    \subfigure[skill-word correlation map with code reinitialization]{
        \includegraphics[width=0.99\linewidth]{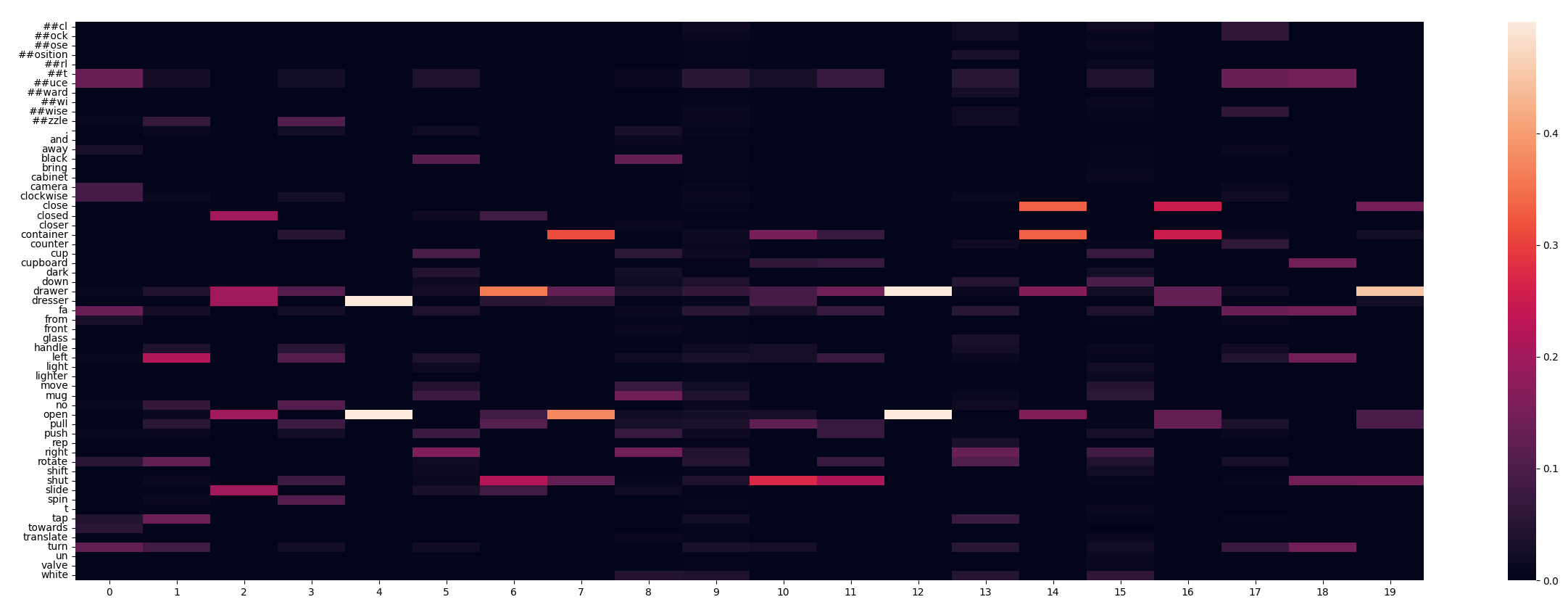}
        \label{fig:lorel_lisa_re}
    }
    \caption{Skill Map on DT model with LISA or LCSD skill learning method.}
    \label{fig:lisa_word}
\end{figure}

We analyze the effectiveness of our skill learning method in the context of Decision Transformer (DT) by comparing it with the original DT-based skill learning model, LISA. In Figure \ref{fig:lisa_word}, we show the skill correlation map for the original LISA model, which only trained on two skills during VQ quantization. However, we can select more skills with our code reinitialization method, as shown in Figure \ref{fig:lorel_lisa_re}. This demonstrates the generalizability of our skill-learning method to other imitation models.

To more specifically describe the meaning of the skill map, Calvin's skill map is analyzed in detail in Figure 7. There are six high-frequency words selected in code 2 that combine to form a single skill language: Push the sliding door to the left. In code 18, 
we can see a combination of multiple expressions of a task: pull/open the drawer/handle. This phenomenon shows that the skill learning of LCSD can fully understand the potential association of human semantics and then better generalize to more tasks.

\begin{figure}
  \centering
  \includegraphics[width=1\linewidth]{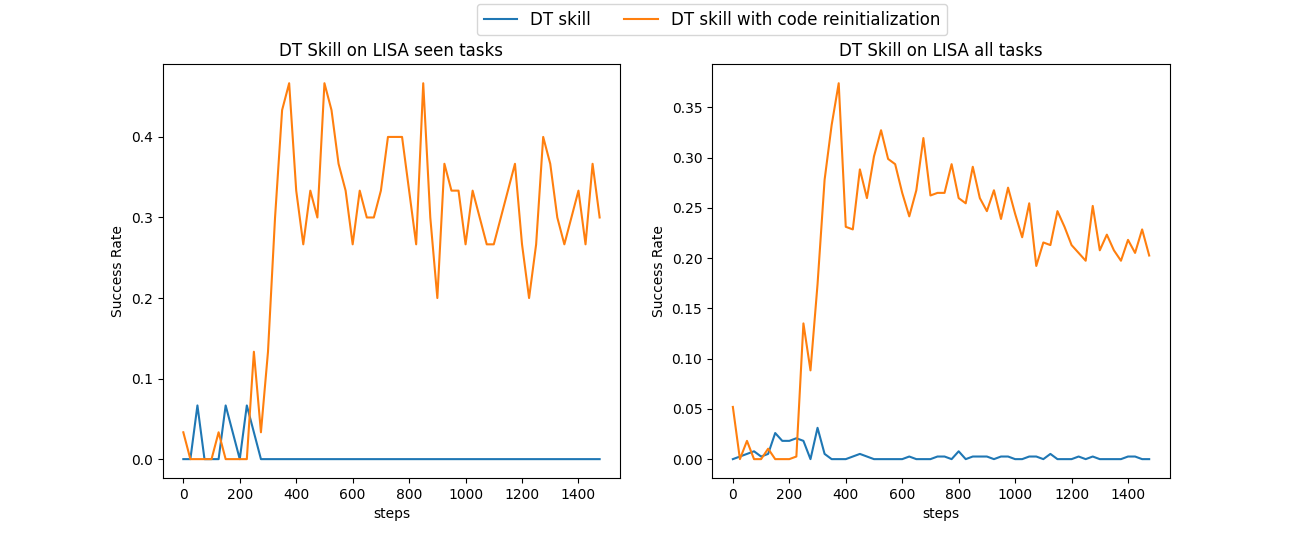}
  \vspace{-5mm}
  \caption{DT model with code reinitialization success rate on LORel.}
  \vspace{-5mm}
  \label{fig:lisa_success}
\end{figure}

In Figure \ref{fig:lisa_success}, we compare the performance of the DT model with and without code reinitialization. We find that the original DT model failed to achieve practical training in almost all task settings because skill learning is limited to index collapse. On the other hand, the advantage of skill dispersion with code reinitialization is reflected in the test accuracy.

In summary, our skill-learning method is effective not only for our LCSD algorithm but also for other imitation models, such as DT-based models. The results suggest that our code reinitialization method can help to generate discrete meaningful skills, leading to better test accuracy and improved skill learning.

\begin{figure}
	\centering
    \subfigure[Diffusion LISA skill map on 30 codes]{
        \includegraphics[width=0.99\linewidth]{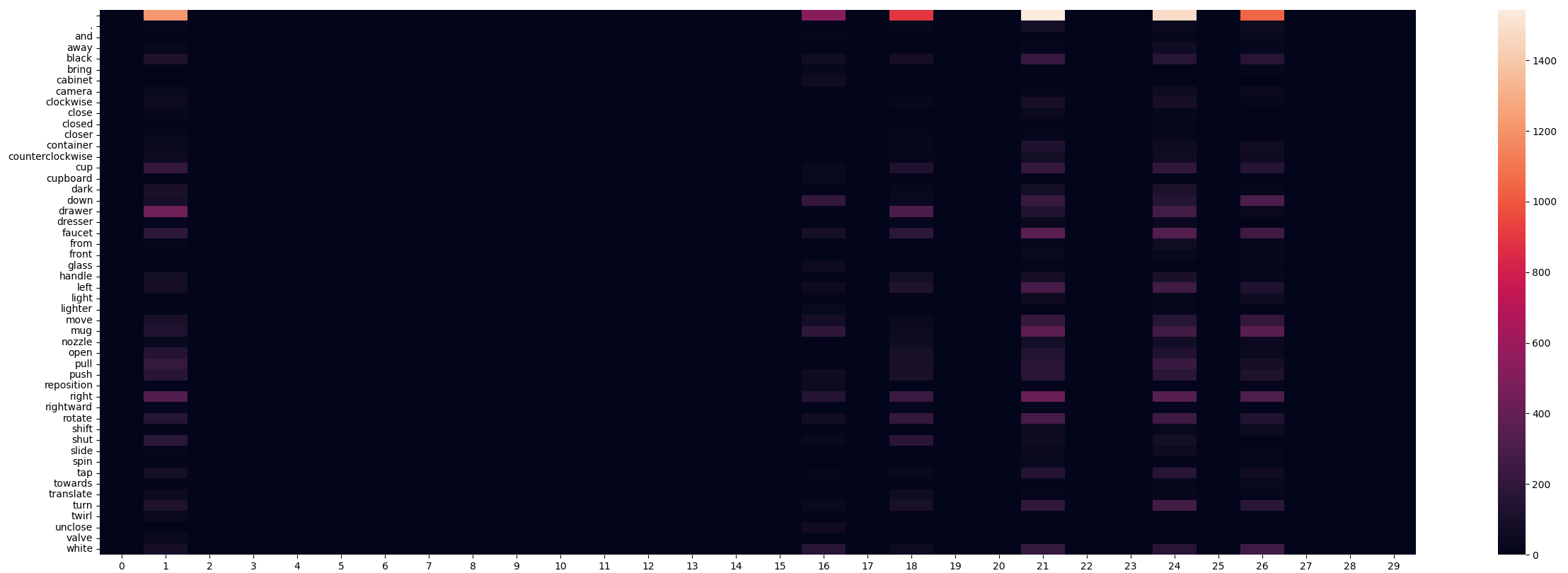}
        \label{fig:diffusion30}
    }
    
    \subfigure[Diffusion LISA skill map on 40 codes]{
        \includegraphics[width=0.99\linewidth]{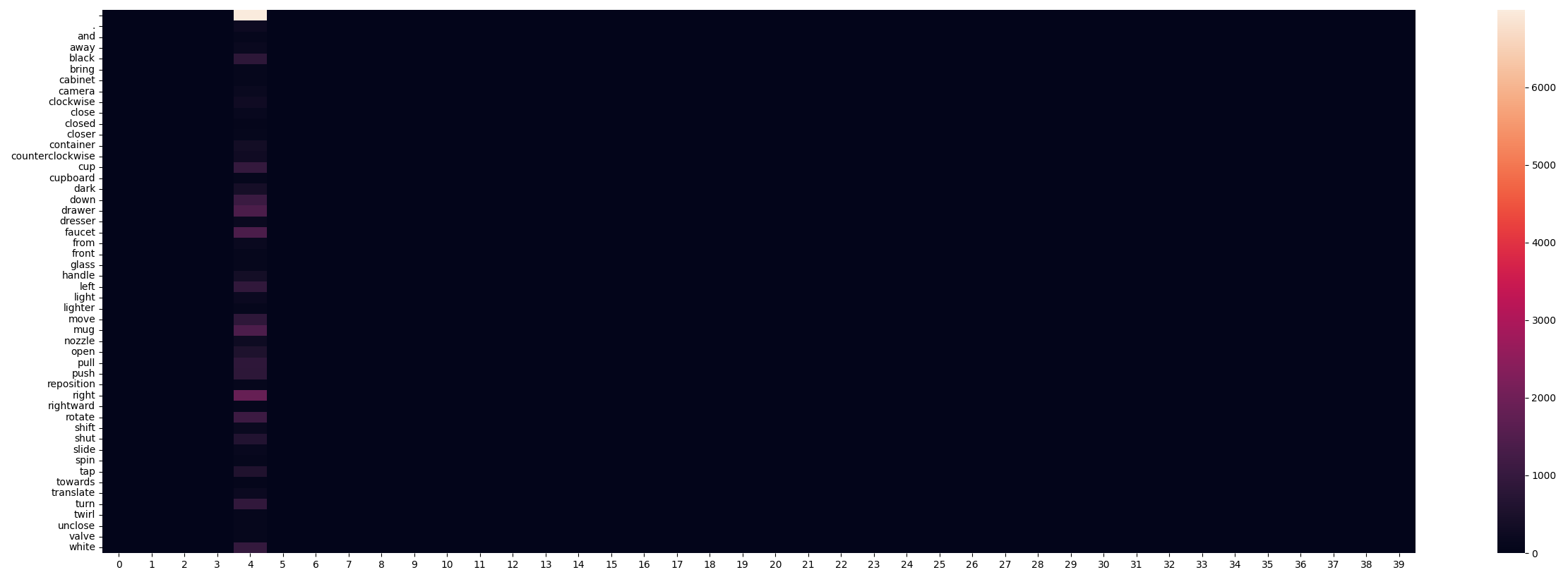}
        \label{fig:diffusiond40}
    }
    \caption{Diffusion LISA Skill map in LORel state environment.}
    \label{fig:diffusion_multi}
\end{figure}

\subsubsection{Different skill map in Varying codebook Size:}
To test the robustness of LCSD, we use the skill-language correspondence graph to identify whether different codebook sizes influence skill learning. For comparison, we also conduct experiments using a basic skill-encoder-only diffusion model, namely the Diffusion Encoder skill model(Abbreviated as Diffusion encoder in Figures). Firstly, we plot our basic diffusion encoder skill model(Diffusion with LISA) in Figure \ref{fig:diffusion_multi} on the codebook with sizes of 30 and 40. From the figures, we observe that skill learning stacked in index collapse, which was more serious when using a larger codebook.

\begin{figure}
	\centering
    \subfigure[LCSD skill map with 30 codes]{
        \includegraphics[width=0.99\linewidth]{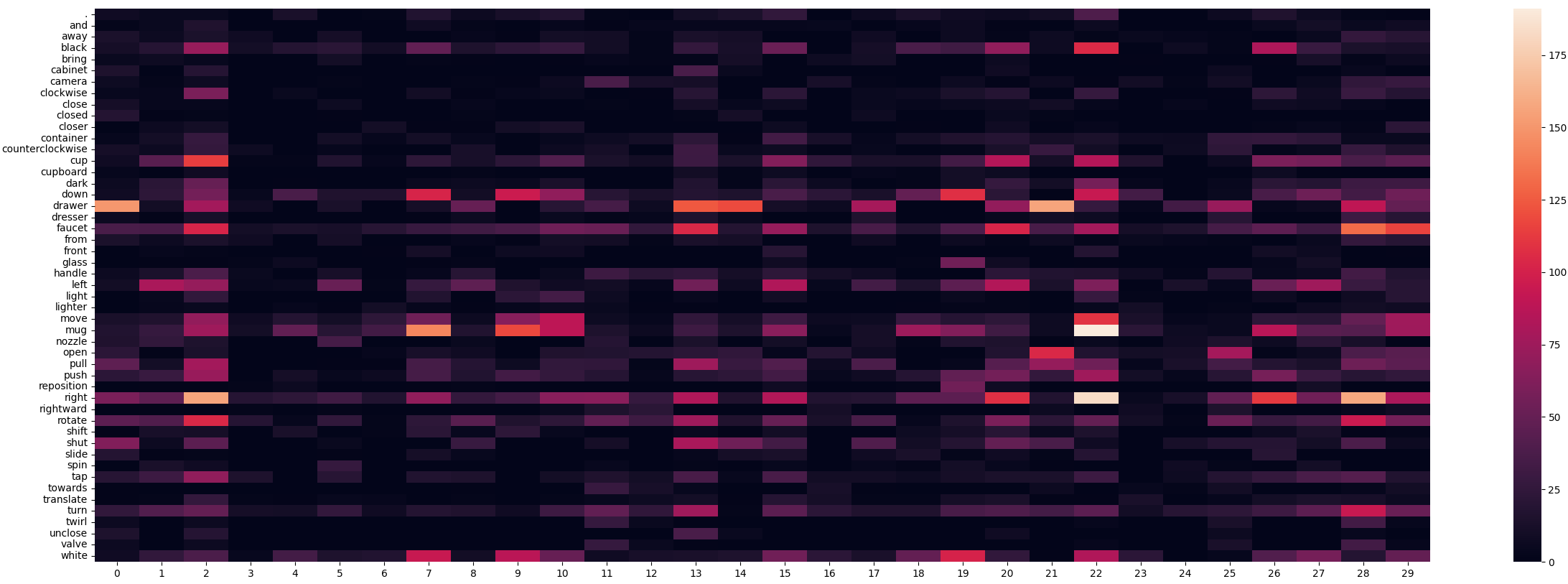}
        \label{fig:lcsd30}
    }
    
    \subfigure[LCSD skill map with 40 codes]{
        \includegraphics[width=0.99\linewidth]{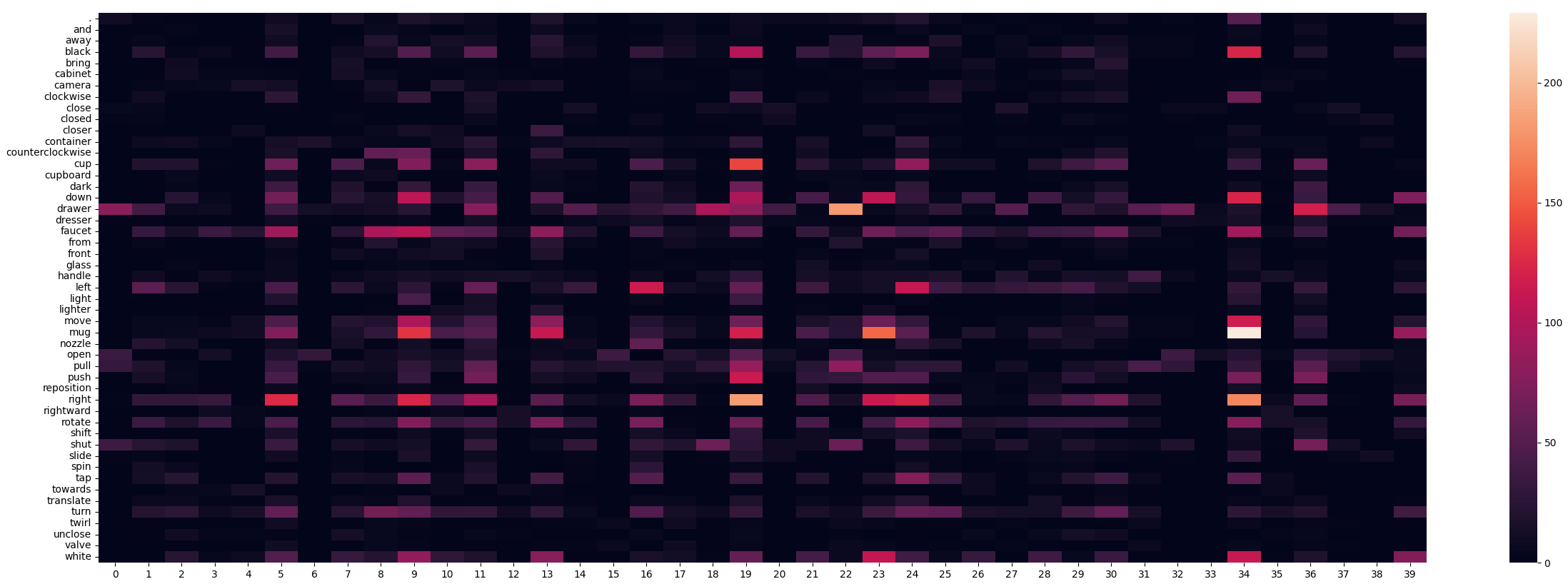}
        \label{fig:lcsd40}
    }
    \caption{LCSD Skill map in LORel state environment.}
    \label{fig:lcsdmulti}
\end{figure}

Next, we test our LCSD algorithm with the same number of skills in LORel, as shown in Figure \ref{fig:lcsdmulti}. We observe that LCSD can learn diverse and precise skills regardless of the codebook size, demonstrating the robustness of our algorithm.

\subsubsection{Skill Maps in Different Environment:}
This section presents our skill correlation maps on the BabyAI and Calvin environments. In the first two tasks of BabyAI, GoToSeq, and SynthSeq, the skills are diverse enough as the tasks are relatively simple to handle. We plot the skill maps for the BabyAI BossLevel tasks to demonstrate the performance of different skill-learning methods in complex language-conditioned tasks, as shown in Figure \ref{fig:babyai_skill}. In contrast, the skills learned by the diffusion encoder are still limited to a small portion of the codebook, and index collapse will be alleviated after adding a language decoder, as shown in Figure \ref{fig:babyai_DE} and Figure \ref{fig:babyai_DED}. It is evident that LCSD generates better discrete and interpretable skills among all the methods in Figure \ref{fig:babyai_LCSD}.

\begin{figure}
	\centering
    \subfigure[Diffusion Encoder skill map with 20 skills]{
        \includegraphics[width=0.99\linewidth]{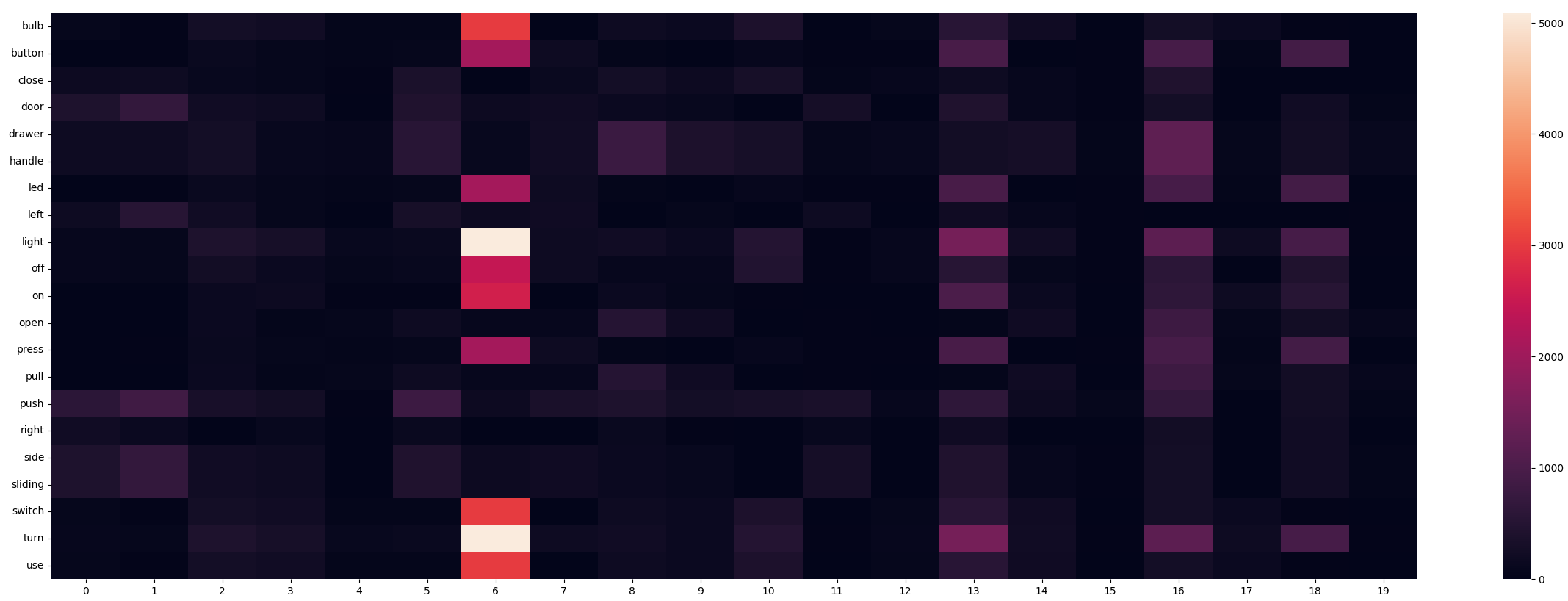}
        \label{fig:calvin_diffusion_20}
    }
    
    \subfigure[LCSD skill map with 20 skills]{
        \includegraphics[width=0.99\linewidth]{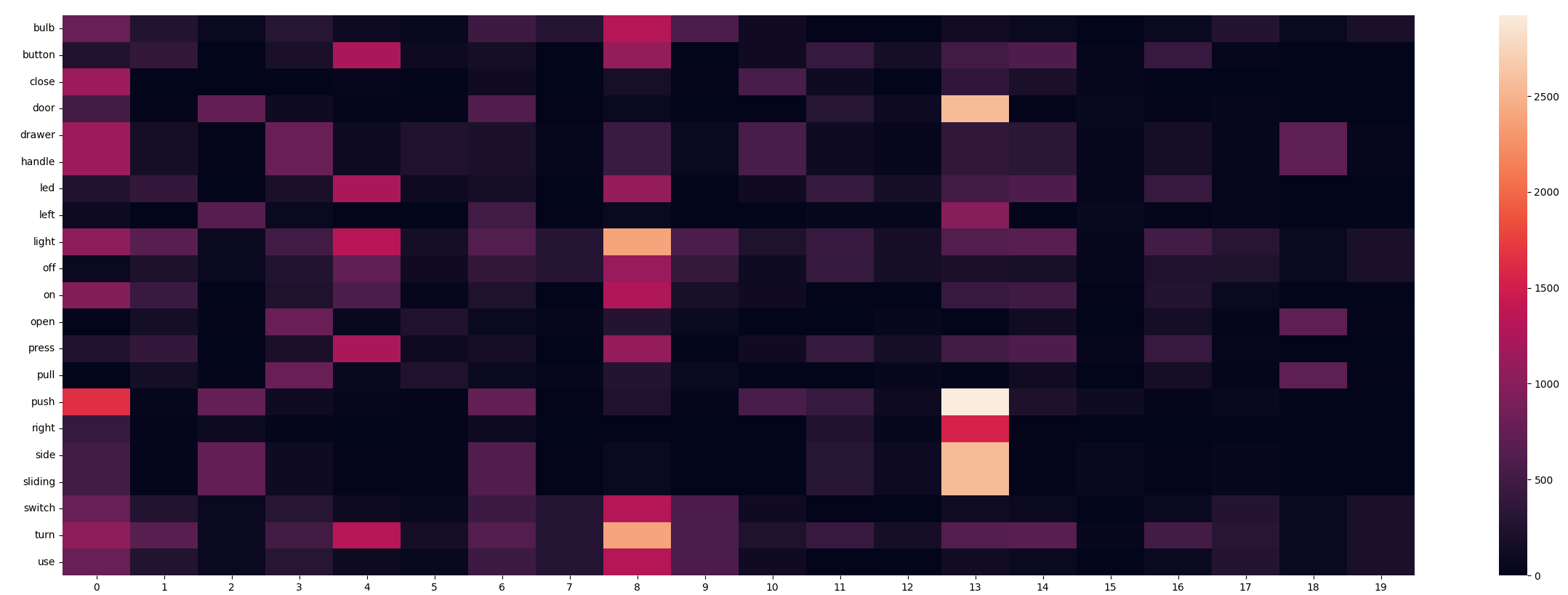}
        \label{fig:calvin_LCSD_20}
    }
    \caption{Skill map in Calvin on codebook length of 20}
 \label{fig:calvin_20}
\end{figure}

\begin{figure}
	\centering
    \subfigure[Diffusion Encoder skill map with 30 skills]{
        \includegraphics[width=0.99\linewidth]{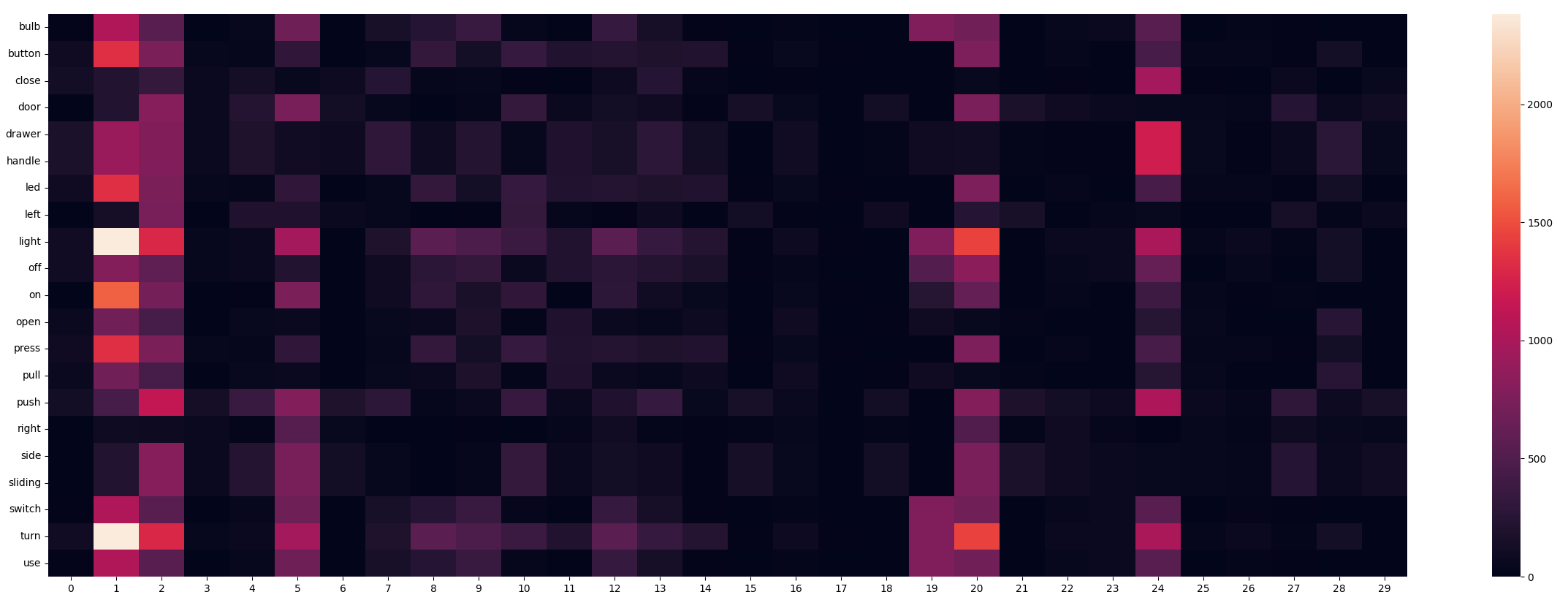}
        \label{fig:calvin_diffusion_30}
    }
    
    \subfigure[LCSD skill map with 30 skills]{
        \includegraphics[width=0.99\linewidth]{Appendix_Figure/calvin_LCSD_30_2.png}
        \label{fig:calvin_LCSD_30}
    }
    \caption{Skill map in Calvin on codebook length of 30}
 \label{fig:calvin_30}
\end{figure}

In the Calvin environment, we find that at a codebook size of 20, the skills learned by the diffusion encoder are close in dispersion to the skills generated by LCSD(Comparing two panels of Figure \ref{fig:calvin_20}).  However, as the number of skills increased, the codebook fell into a small subset of skill vector during skill learning, as shown in Figure \ref{fig:calvin_diffusion_30}.  In contrast, LCSD is able to maintain skill diversity even with a larger codebook, which corresponds to our stable performance on different tasks (Figure \ref{fig:calvin_LCSD_30}).


In summary, our skill correlation maps in different environments demonstrate the effectiveness of our LCSD algorithm in learning diverse and precise skills, even in complex language-conditioned tasks. The ability of our algorithm to maintain skill diversity regardless of the codebook size and number of reconstruction options is a crucial advantage, enabling us to handle large and complex datasets with varying hyperparameters.

\begin{figure*}
	\centering
    \includegraphics[width=1\linewidth]{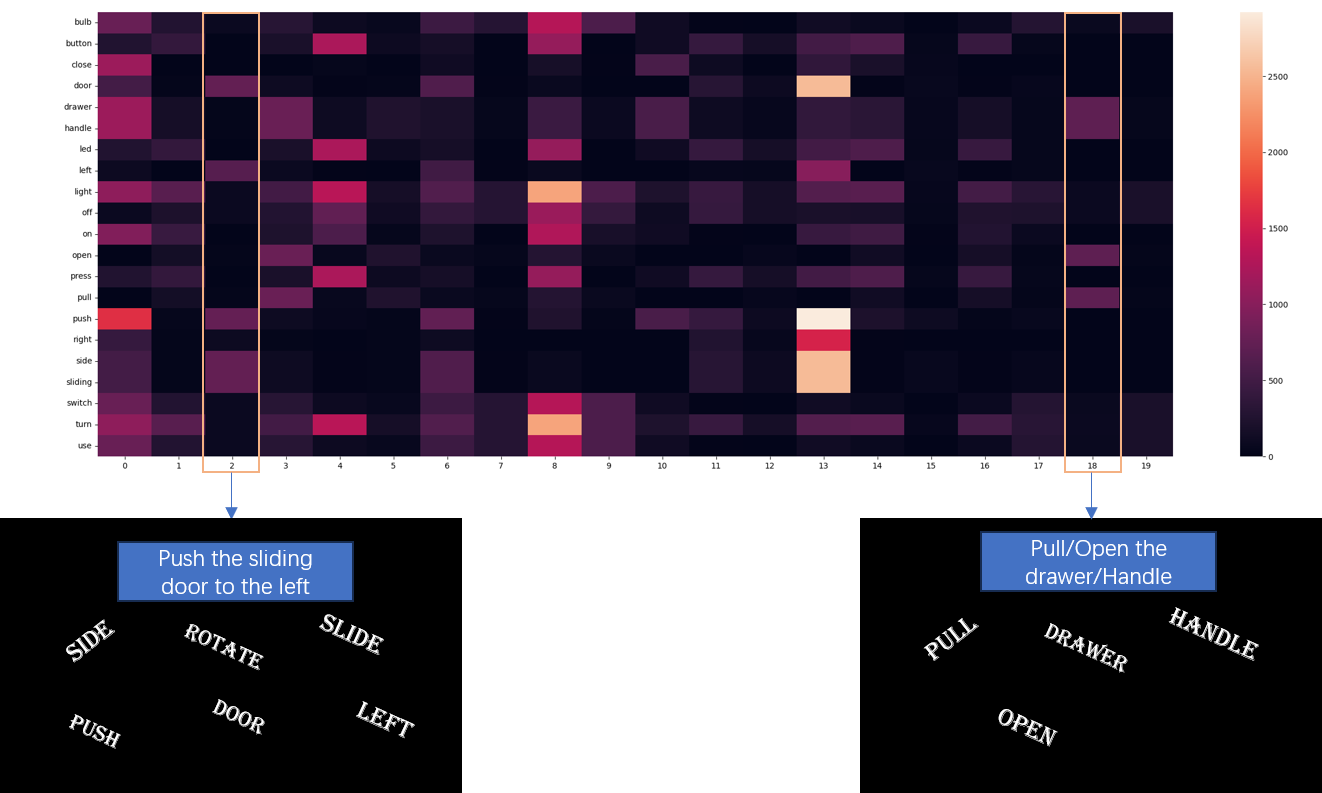}
    \caption{\textbf{Explanation of certain skill codes in CALVIN}. Skill maps show how our skills are correlated with words in evaluation. As can be seen from the second column, six words are chosen and can be formed into a sentence: Push the Sliding door to the left. Also, in code 18, this skill corresponds to the task "open drawer" and can recognize different representations: open/pull, drawer/handle. }
    \label{fig:exp}
\end{figure*}

\begin{figure*}
	\centering
    \subfigure[LCSD skill map]{
        \includegraphics[width=0.49\linewidth]{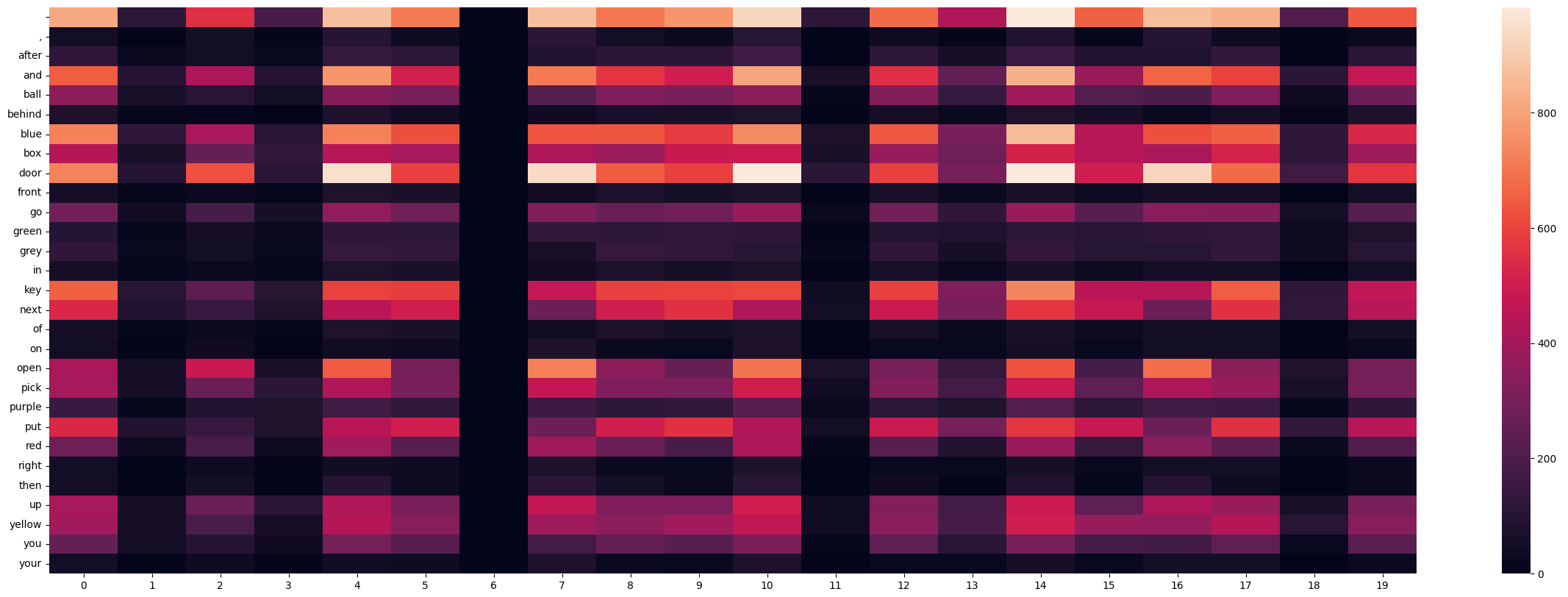}
        \label{fig:babyai_LCSD}
    }

    \subfigure[Diffusion LISA skill map]{
        \includegraphics[width=0.45\linewidth]{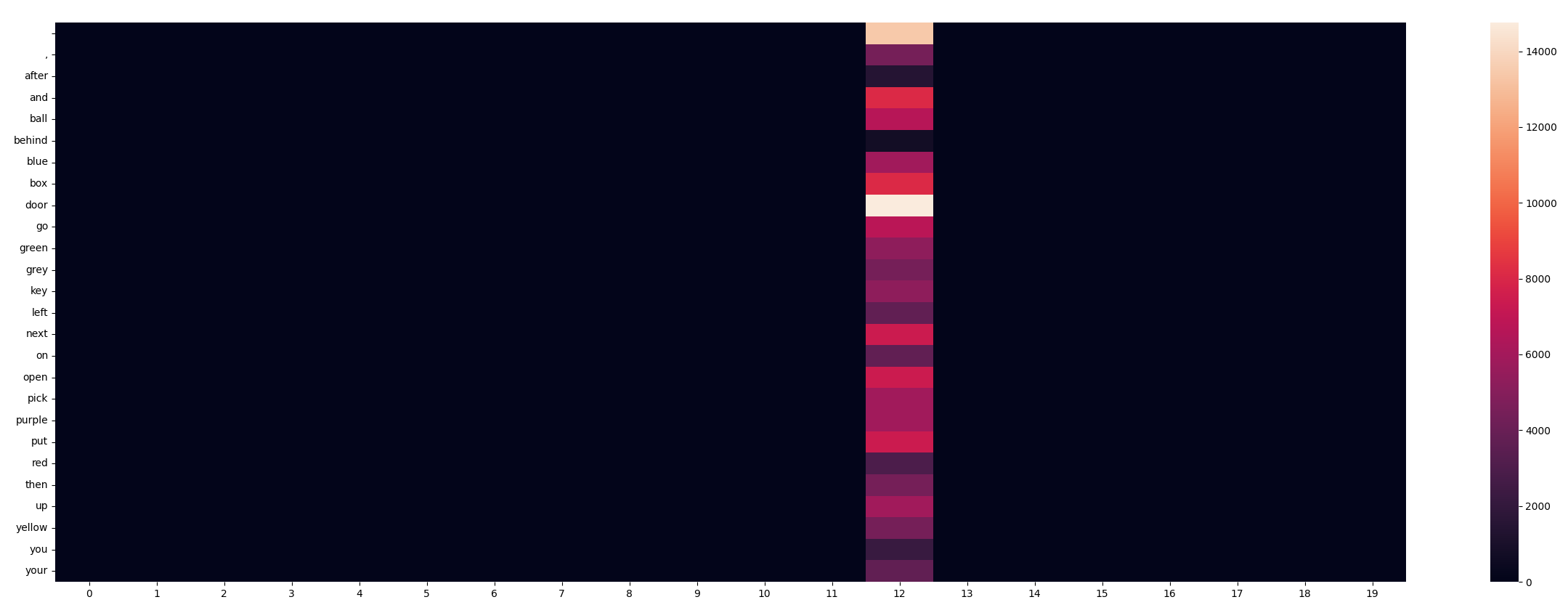}
        \label{fig:babyai_DE}
    }
    \subfigure[Diffusion LISA skill map with reinit]{
        \includegraphics[width=0.45\linewidth]{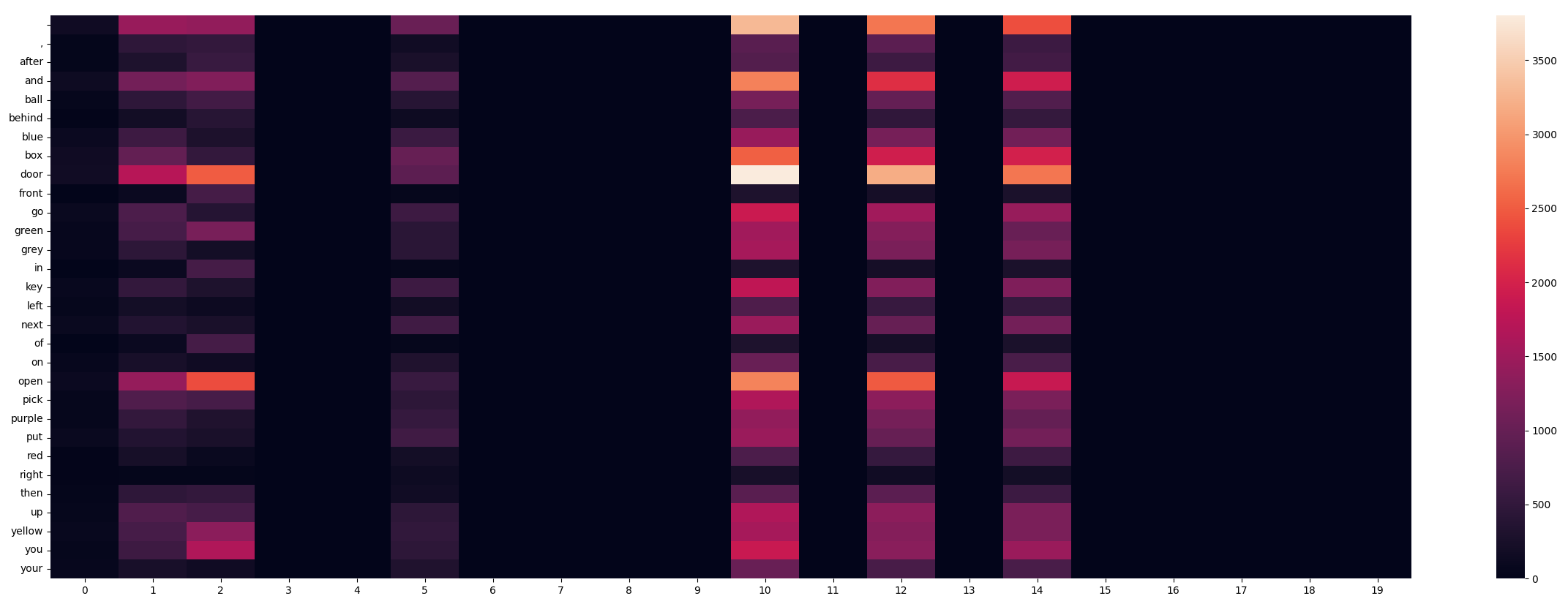}
        \label{fig:babyai_DED}
    }
 \caption{Skill map in BabyAI Bosslevel}
 \label{fig:babyai_skill}
\end{figure*}

\end{document}